\documentclass[10pt,twocolumn,letterpaper]{article}

 \usepackage{cvpr}              
\usepackage{graphicx}%
\usepackage{multirow}%
\usepackage{amsmath,amssymb,amsfonts}%
\usepackage{amsthm}%
\usepackage{mathrsfs}%
\usepackage[title]{appendix}%
\usepackage{xcolor}%
\usepackage{textcomp}%
\usepackage{manyfoot}%
\usepackage{booktabs}%
\usepackage{algorithm}%
\usepackage{algorithmicx}%
\usepackage{algpseudocode}%
\usepackage{listings}%
\usepackage{tabularx}
\usepackage{calc}
\usepackage{adjustbox,lipsum}
\usepackage{natbib}
\usepackage[utf8]{inputenc} 
\usepackage[T1]{fontenc}    
\usepackage{url}            
\usepackage{booktabs}  
\usepackage{mathtools}
\definecolor{cvprblue}{rgb}{0.21,0.49,0.74}
\usepackage[pagebackref,breaklinks,colorlinks,allcolors=cvprblue]{hyperref}

\def\paperID{10485} 
\def\confName{CVPR}
\def\confYear{2025}

\title{Potential Field Based Deep Metric Learning}

\author{
Shubhang Bhatnagar \quad \quad Narendra Ahuja\\
University of Illinois Urbana-Champaign\\
{\tt\small \{sb56, n-ahuja\}@illinois.edu}
}

\begin{document}
\maketitle
\begin{abstract}
Deep metric learning (DML) involves training a network to learn a semantically meaningful representation space. Many current approaches mine n-tuples of examples and model interactions within each tuplets. We present a novel, compositional DML model that instead of in tuples, represents the influence of each example (embedding) by a continuous potential field, and superposes the fields to obtain their combined global potential field. We use attractive/repulsive potential fields to represent interactions among embeddings from images of the same/different classes. Contrary to typical learning methods, where mutual influence of samples is proportional to their distance, we enforce reduction in such influence with distance, leading to a decaying field. We show that such decay helps improve performance on real world datasets with large intra-class variations and label noise. Like other proxy-based methods, we also use proxies to succinctly represent sub-populations of examples. We evaluate our method on three standard DML benchmarks- Cars-196, CUB-200-2011, and SOP datasets where it outperforms state-of-the-art baselines.
\end{abstract}
\section{Introduction}
\label{sec:intro}
The goal of visual metric learning is to learn a representation function where common distance metrics like the $\ell_{2}$ distance capture similarity between images. Semantic similarity is useful for applications such as image/video retrieval and search \cite{video_search,proxy_nca, proxy_anchor, bhatnagar2023piecewiselinear}, open set classification and segmentation \cite{open_set_activity,dml_open_semantic, open_set_robotics, rawlekar2025improving}, few-shot learning \cite{prototypical, LearningTC_few_shot}, person re-identification\cite{Chen_2017_CVPR,xiao2017joint} and face verification \cite{face_verfication_inthewild,sphereface,arcface}. Deep neural networks have helped make significant advances in visual metric learning \cite{schroff_facenet_2015}, as they are trained to learn a non-linear mapping of images onto a semantic space. 
 
\textbf{Existing Methods:}  Popular loss functions to train such networks help force tuplets of examples/samples/embeddings of the same class to be closer than those in different classes. Examples include the triplet loss \cite{ triplet_orig}, the contrastive loss \cite{chopra2005learning} and their extensions \cite{sohn2016improved,multisimilarity_dml}.

The utility of supervision provided by samples in a tuplet depends on samples’ relative locations and labels. Most tuplet-based methods use sample mining strategies \cite{schroff_facenet_2015,harwood_smart_2017} to choose a subset of informative tuplets to train the network, without which they suffer from lower performance and slower convergence \cite{wu_sampling_2018}. Such strategies not only require the computationally demanding calculation of the relative hardness of all tuples formed (O($n^{3}$) for triplets among $n$ samples), but also incur a loss of valuable supervision by discarding interactions. Specifically, limiting interactions to only a subset of all possible tuplets leads to greater susceptibility to noise \cite{Levi_2021_ICCV}, overfitting on the training set \cite{context_icml_2023}, and reduced learning of intra-class variations and generalizable features \cite{revisiting_dml}.

Proxy-based methods, first proposed in \cite{proxy_nca} attempt to mitigate some of these shortcomings by instead using class representative ‘proxies’ as one or more hypothesized centers for each class.  Computation of the loss is then in terms of sample-proxy distances instead of sample-sample distances, helping circumvent the problem of tuplet sampling.  However, these methods have been found to learn less generalizable intra-class features due to the exclusive use of proxy-sample distances while ignoring sample-sample distances\cite{roth2022non}. This effect is compounded when proxies, whose locations are themselves learned, end up far from sample embedding locations they are intended to represent.

 \textbf{Proposed Representation:}  To address these challenges, we propose Potential Field based Deep Metric learning (PFML), a metric learning framework that uses a continuous, potential field representation of all data samples motivated by potential fields used to model influence of charges in physics. Besides its analogical relevance, our motivation for using this representation comes from its previous, very successful uses, where it also helps overcome complexity in modeling interactions. These applications include planning paths of a robot avoiding obstacles \cite{potential_path_main} and planning an efficient strategy for a robot manipulator to grasp an object \cite{khatib_potential}. 
 
 In PFML, we view each data sample as a charge, creating a field. The fields due to the individual samples are added (superposed) to obtain the global potential field.  This global field thus constitutes a \emph{compositional} representation, capturing the combined influence at a given location (or sample) as a weighted \emph{addition} of the influences of the individual samples. Each sample generates two types of fields: (1) an attraction field to drive similar examples closer to it, and (2) a repulsion field that drives dissimilar examples away from it, each weakening with distance.  

We also make use of proxies like previous work, but only to augment the potential field generated by the current batch of samples. This is in contrast to past work that completely replaces sample-sample interactions with sample-proxy interactions. Preserving sample-sample relations through the potential field helps learn better intra-class features.

\textbf{Advantages of Potential Field:} (1) Using the potential field representation enables us to model interactions among all sample embeddings, as opposed to modeling those among smaller number of samples (or proxies) as done in methods using point-tuplet based (e.g., triplet) loss. (2) Addition of potentials due to all points increases robustness to noise since the effect of noise on interactions among a smaller number of samples will have a larger variance while also (3) Considering all interactions improves the quality of features learned, unlike tuplet based methods which lose performance when all interactions are considered\cite{wu_sampling_2018}. (4) This is because of
a major feature that differentiates our potential field from past methods: The variation of strength of interaction between two points as the distance between them increases: instead of remaining \emph{constant} or even becoming \emph{stronger}, as is the case with most existing methods, e.g. \cite{chopra2005learning, proxy_anchor}, in our model it becomes \emph{weaker} with distance.\footnote{It remains constant over a small radius and decays outside; however, for brevity, we will disregard this detail in the rest of this paper, and refer to the influence as decaying with distance, or as the decay property.} 
The decay property is helpful in several ways: (4a) It ensures the intuitive expectation that two distant positive samples are too different to be considered as variants of each other, helping treat them as different varieties (e.g., associated with different proxies). (4b) The decay property also significantly improves PFML performance for the specific type of noise affecting labels, e.g., due to annotation errors common in real-world datasets (Sec. \ref{sec:noise_exp}). (4c) As a result of the decay, the learned proxies remain closer to (at smaller Wasserstein distance $W_{2}$ from) the sample embeddings they represent than for (e.g., current proxy-based) methods where interactions strengthen with distance \cite{proxy_anchor,proxy_nca, soft_triplet}; Sec \ref{sec:discussion}), thereby enhancing their desired role.


\textbf{Contributions:}  The main contributions of the proposed approach are:
 \begin{itemize}
\item We present PFML, a novel framework for metric learning with a compositional representation using a continuous, potential field to model all sample interactions instead of only interactions between a subset of all possible n-tuplets. This enables better feature learning, especially in the presence of annotation noise.
\item PFML reverses the almost universally used model of inter-sample interactions strengthening with distance; interaction strength instead  weakens with distance.  This (1) helps improve robustness in the presence of label noise, and (2) we prove that it helps better align the distribution of proxies with the sample embeddings they represent. 

 \item We evaluate PFML on three zero-shot image retrieval benchmarks: Cars-196, CUB-200-2011, and the SOP dataset where it outperforms current, state-of-the-art techniques in both the standard no-label-noise scenario, and with even greater margins (>7\% R@1) in the more realistic cases of label noise.
 \end{itemize}
\section{Related Work} \label{sec:rel_work}
 Broadly, DML may be divided into two types, ranking/pair-based and proxy-based \cite{proxy_anchor, multisimilarity_dml}. 

\textbf{Ranking/Pair based DML}: These methods utilize losses 
that derive information from relations between inter-sample distances in pairs/ tuples of sample embeddings.
Examples include the classical contrastive loss \cite{chopra2005learning} or the triplet loss\cite{triplet_orig, wu_sampling_2018, schroff_facenet_2015}, which extends the concept to take into account relative distances in a triplet. \cite{sohn2016improved} and \cite{sopdataset} generalize these constraints to include multiple dissimilar samples. 
Mining informative pairs/tuples from a mini-batch plays an important role in the effectiveness of pair-based methods. Several strategies \cite{schroff_facenet_2015, yuan_hard-aware_2017, harwood_smart_2017} have been proposed to pick samples based on their relative difficulty, with hard and semi-hard negative (picking pair of inter-class examples close together) and easy positive (same class examples close together) \cite{xuan_easy, Levi_2021_ICCV}  being popular. PFML naturally emphasizes such local (dis) similarities by virtue of its decaying formulation without the complexity of mining tuplets. Also note that unlike PFML, none of these strategies optimize all interactions among a batch of samples.

\textbf{Proxy-based DML}:  ProxyNCA \cite{proxy_nca} and its variants \cite{probab_proxy, proxy_ncapp} attempt to minimize the distance of a sample from the proxy of its class while maximizing the distance from proxies of other classes. 
Proxy Anchor \cite{proxy_anchor} builds on these by proposing a modified loss that takes into account sample hardness. \cite{soft_triplet, arcface} introduced the use of multiple proxies per class to better model sub-clusters. 

\textbf{Potential Field}: Potential fields have been used in a variety of applications to model interactions of objects. Of interest here is their use in robot path planning \cite{potential_path_main, potential_path2}, shape representation \cite{potential2,potential1}, image segmentation \cite{ahuja1996transform,tabb1997multiscale} and object manipulation \cite{khatib_potential}. Here, they have been used to direct a robot towards its goal using an attraction field while avoiding obstacles using repulsion fields. Training supervision for DML shares a similar goal of moving sample embeddings towards other nearby samples of the same class, while avoiding dissimilar examples belonging to other classes.

    
     

\begin{figure}
\centering
    \centering
        \includegraphics[width=0.48\textwidth]{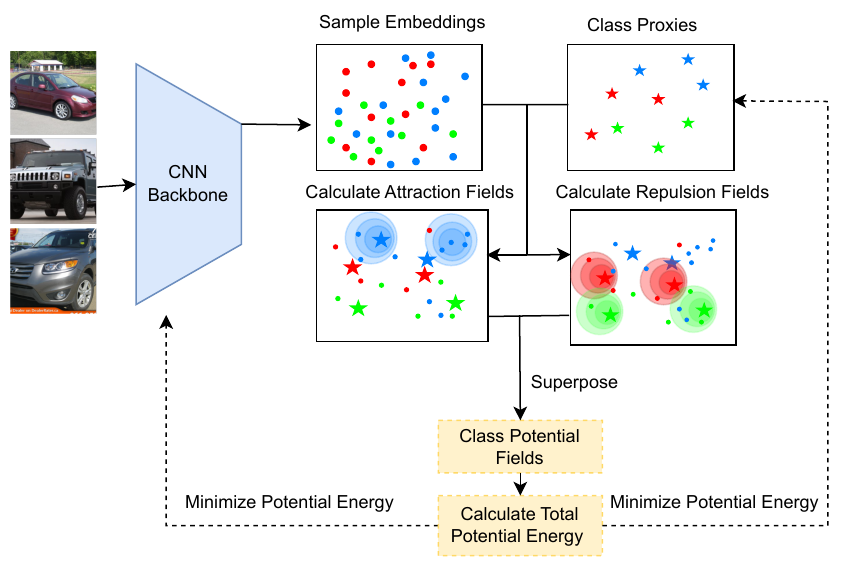}
      \caption{Overview of our Potential-field based DML pipeline. The process includes (1) Computing attraction and repulsion fields generated by each embedding and proxy, (2) Computing the class potential fields by superposition of individual fields (3) Evaluating total potential energy by summing up the potentials of embeddings and proxies under the class potential field and (4) Updating locations of sample embeddings (through network parameters) and proxies to minimize total potential energy through backprop.}
    \label{fig:flow_summary}
    \vspace{-15pt}
\end{figure}

 \section{Method} \label{sec:method}
 \subsection{Problem Setup}
The goal of deep metric learning is to define a semantic distance metric $ d(\textbf{x}_{1}, \textbf{x}_{2})$ for any pair of samples $\textbf{x}_{1}, \textbf{x}_{2} \in \mathcal{D}$, where $ \mathcal{D}$ $ = \{ \textbf{x}_{i}, y_{i} \}, i \in \{1 .... \lvert \mathcal{D}\rvert \} $ and labels $y_{i} \in \{1,\hdots N\}$ represent N classes. The learned distance metric
for any two points $ d(\textbf{x}_{1}, \textbf{x}_{2})$ should represent their semantic dissimilarity.
Instead of directly learning the distance metric, a projector function $f_{\theta}$ is often represented by a neural network, parameterized by $\theta$ learned from data and distance $ d(\textbf{x}_{1}, \textbf{x}_{2}) = \|f_{\theta}(\textbf{x}_{1}),f_{\theta}(\textbf{x}_{2})\|_{2}$ is given by the Euclidean distance between the projections/embeddings. For simplicity, we will refer to the embeddings $f_{\theta}(\textbf{x}_{i})$ belonging to a batch $\mathcal{B}$ sampled from $\mathcal{D}$ as $\textbf{z}_{i}$.

\subsection{The Potential Field }
 With our potential field representation, embeddings need to be driven towards other nearby embeddings belonging to the same class, while also being driven away from embeddings of other classes. This is reminiscent of the behavior of an isolated system of electric charges, where dissimilar charges are drawn together while similar ones are repelled. Analogously, we define attractive and repulsive potentials created by a single embedding, with the net potential generated by a set of samples given by superposing individual potentials exerted by each embedding. The gradient of the net potential at a point yields the net force (on network update) on a sample embedding located at that point.


\begin{figure*}
        \centering
    \includegraphics[width=0.9\linewidth]{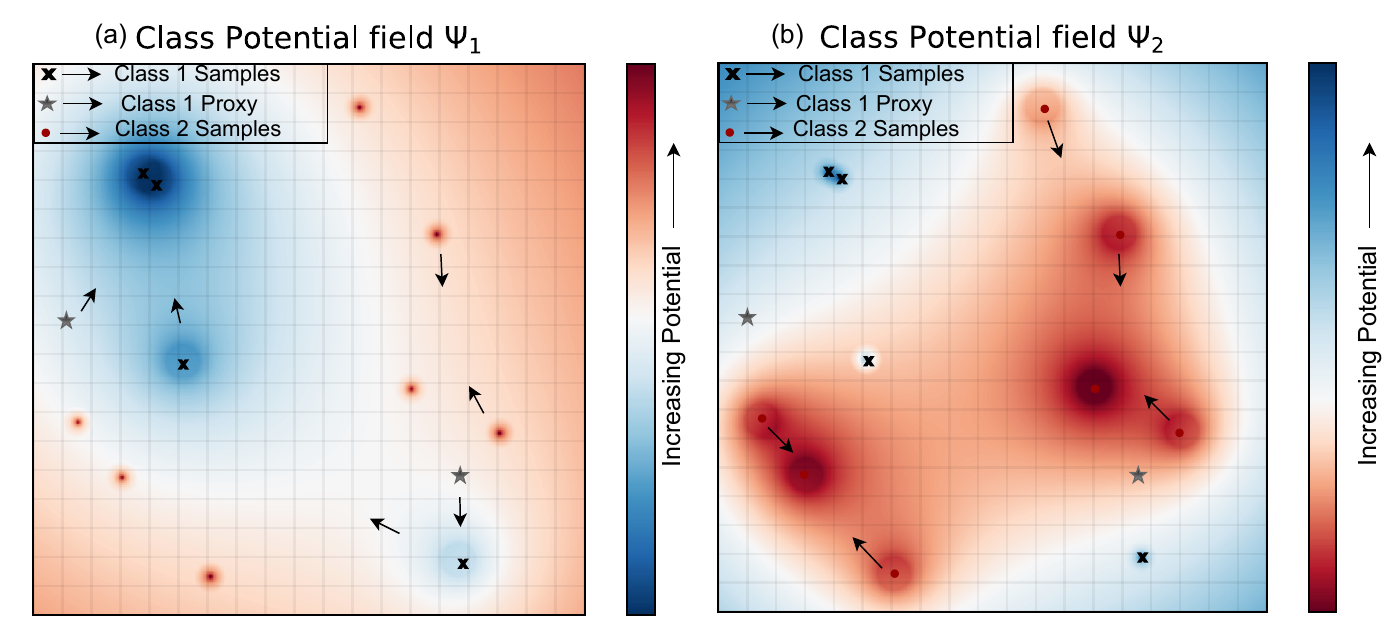}
    \caption{An example of the class potential fields $\Psi_{1}$ and $\Psi_{2}$ (Sec. \ref{sec:potential_combined}), created by superposing the fields of individual embeddings (Sec \ref{sec:attraction}) belonging to Classes 1 and 2. Arrows denote the gradient of the respective potentials representing the net force on them. (a) $\Psi_{1}$ draws samples/proxies of class 1 towards nearby samples/ proxies of class 1 while keeping them at least $\delta$ distance away from embeddings of class 2. Proxies at starred locations are drawn towards the nearest Class 1 embeddings which are potential minima, helping them better model the data distribution (Sec 3.4). (b) $\Psi_{2}$ draws embeddings of class 2 towards other nearby embeddings of class 2 instead of distant class 2 embeddings, which might be a significantly different variant of the class (or potentially a mislabeled data point), helping better feature learning (or improving noise robustness if mislabeled). $\Psi_{2}$ also keeps class 2 embeddings $\delta$ distance them away from embeddings of class 1. }
    \label{fig:field_fig}
    \vspace{-15pt}
\end{figure*}
As samples of each class have a different set of similar and dissimilar points, we define a separate potential field $\Psi_{j},$ for each class $\  j \in \{1,\hdots N\}$.
The class potential field $\Psi_{j}$ models the force applied on embeddings belonging to class $j$ due to all embeddings. It is formed from a superposition of potentials belonging to sample embeddings from all classes, with embeddings from the same class exerting an attractive potential field (to pull similar points closer) while those from other classes exert a repulsive potential field (to push apart dissimilar points). We assume that each sample embedding $z_{i}$ has a unit charge, so the net force on it is the gradient of the potential field $\Psi_{y_{i}}$ of its class.
We design the attractive and repulsive potential functions by taking motivation from both, electrostatic fields and metric learning literature, which we discuss in detail in the following subsections. Figure \ref{fig:flow_summary} provides an overview of our complete pipeline of generation and use of the potential field for metric learning .


\vspace{-10pt}
 \subsubsection{Potential field exerted by a single point}\label{sec:attraction}
By using forces on the embeddings, we wish the training process to move the embeddings to a state of equilibrium at the end of training in which all embeddings from a class are nearby. To have a potential function exhibiting this behavior, we ensure that individual potential field has the following two properties: \\
\textbf{Property 1 - Extremum Property:} Potential generated by an individual embedding achieves an extremum at the embedding (minimum for attraction field, maximum for repulsion). This ensures that similar points are pulled together and dissimilar ones are pushed apart on minimizing potential. \\
\textbf{Property 2 - Decay Property:} The magnitude of the gradient of the potential (force) due to an embedding decays with distance. Such a decay
ensures two desired characteristics. (1) An embedding is not pulled towards a distant sample embedding belonging to the same class, which is likely to be too different to be considered as a variant of it; instead, the decayed influence helps treat the distant sample as significantly different variety of the class, associated with other nearby embeddings of the class. (2) Like for a single charge in the Extremum property, a superposition of potentials of different embeddings still yields local minima at or near sample embeddings.
As shown in Sec \ref{sec:discussion}, this helps proxies come close to such minima and thus help better model the data distribution.
\\
Given these properties motivated by electrostatic fields, we select the attraction and repulsion potentials $\psi_{att}$, $\psi_{rep}$ generated by an embedding $\textbf{z}_{i}$, with their strength at location $\textbf{r}$ defined as:
 \begin{align}
     \psi_{att}(\textbf{r}, \textbf{z}_{i}) :&= \begin{dcases} 
    -\dfrac{1}{\delta^{\alpha}} & \textrm{if } \| \textbf{r} - \textbf{z}_{i} \|_{2} < \delta \\
    -\dfrac{1}{\| \textbf{r} - \textbf{z}_{i}  \|_{2}^{\alpha}} & \textrm{otherwise.} 
    \end{dcases} \label{eq:attract_field_point}
    \end{align}
    \begin{align}
    \psi_{rep}(\textbf{r}, \textbf{z}_{i}) :&=\begin{dcases} 
     \dfrac{1}{\| \textbf{r} - \textbf{z}_{i} \|_{2}^{\alpha}} & \| \textbf{r} - \textbf{z}_{i} \|_{2} < \delta  \\
    \dfrac{1}{\delta^{\alpha}} & \textrm{otherwise.} \end{dcases} \label{eq:repel_field_point} 
 \end{align}
    
%
  Here, $\alpha$ is a hyperparameter that determines the rate of decay of the field (we evaluate it's effect in Sec. \ref{sec:ablation} instead of fixing $\alpha = 1$ as done in electrostatic fields). 
  Note that the attraction potential is negative and increases with distance, implying that the force between the embedding and the point is attractive. The repulsion potential is positive and decreasing with distance, indicating repulsion. 

  A constant attraction potential inside the $\delta$ sphere (i.e., force = 0) ensures that sample embeddings within $\delta$ distance of one another do not continue to attract each other, thus preventing the network from embeddings from collapsing to a single point which harms generalizability of the features learned  (analogous to the effect margins used in tuplet-based losses have).  

 Limiting the repulsion potential to a sphere of radius $\delta$ ensures that embeddings not belonging to the same class are driven to a distance $> \delta$. Since the aforementioned constant attraction potential ensures that $\delta$ is the maximum distance between samples $z_{i}$ from the same class, this leads to connected components of $\delta$ spheres formed by nearby embeddings of the same class.
 Extending repulsion to distances $> \delta$ does not provide any useful additional supervisory information while forcing further unmotivated inter-class separation; as we empirically show in Sec. \ref{sec:ablation}, such excessive separation does indeed harm performance.
 \subsubsection{Proxies to Represent Sample Populations} 
To represent the image embeddings not sampled in the current batch $\mathcal{B}$, we propose to use a set of $M$ proxies per class for each of the N classes, $ \textbf{p}_{j,k}, k \in\{1, \hdots M \}, j \in\{1, \hdots N \}$, and augment the potential field generated by the embeddings in the batch by those generated by the proxies. The proxies serve as stand-ins for the larger numbers of out-of-the-batch embeddings for calculating the attraction and repulsion fields, while also being affected by the field themselves like all embeddings. The proxies, being learnable parameters with high learning rates, tend to gravitate toward the nearest class modes of the potential field (and hence the class data distribution). The number of proxies M chosen should be large enough to capture the class-modes.
 
\subsubsection{Class Potential Field}\label{sec:potential_combined}
Given a batch $\mathcal{B}$ of samples and all proxies, we calculate the class potential field $\Psi_{j}(\textbf{r})$ at a location \textbf{r} due to a class $j$ as a superposition of their attraction and repulsion potentials: 
\begin{align}
    \Psi_{j}(\textbf{r}) & =  \Psi_{j,att}(\textbf{r})+ \Psi_{j,rep}(\textbf{r})  \text{, where }\label{eq:class_potential}\end{align}
    \begin{align}
     \Psi_{j,att}(\textbf{r})& =\sum_{i=1}^{\|\mathcal{B} \|, y_{i} = j } \psi_{att}(\textbf{r}, \textbf{z}_{i}) + \sum_{k=1}^{M }  \psi_{att}(\textbf{r}, \textbf{p}_{j,k}) \label{eq:attract_field_class} \end{align}
     \begin{align}
      \Psi_{j,rep}(\textbf{r}) &= \sum_{i=1}^{\|\mathcal{B} \|, y_{i} \neq j } \psi_{rep}(\textbf{r}, \textbf{z}_{i}) + \sum_{\gamma = 1, \gamma \neq j}^{N} \sum_{k=1}^{M } \psi_{rep}(\textbf{r}, \textbf{p}_{\gamma,k})  \label{eq:repel_field_class} 
\end{align}
Figure \ref{fig:field_fig} visualizes the class potential fields $\Psi_{1}$ and $\Psi_{2}$ for a given set of embeddings belonging to two classes.
\subsection{Training}
Using gradient descent, the network is trained to minimize the total potential energy $\mathcal{U}$  of all proxies and embeddings in the batch $ \textbf{z}_{i} \in \mathcal{B}$ by applying a force on embeddings and proxies in the direction of $- \nabla \Psi_{y_{i}}(\textbf{z}_{i})$. $\mathcal{U}$ given by :
\begin{align}
    \mathcal{U} = \sum_{i=1}^{\|\mathcal{B}\|} \Psi_{y_{i}}(\textbf{z}_{i}) +  \sum_{j=1}^{N}\sum_{k=1}^{M} \Psi_{j}(\textbf{p}_{j,k})
    \label{eq:potential_total}
\end{align}

    



\subsection{Why decay interaction strength with distance ?}\label{sec:discussion}
PFML uses a potential field where the gradient (interaction strength) decays with distance, which is unlike previous approaches. This leads to the two benefits described in Sec \ref{sec:attraction}, which we discuss in more detail here.
\subsubsection{Robustness when learning with datasets having large intra-class variations and noisy labels}
The decay property helps PFML model all interactions even when using datasets with large intra-class variations. As seen in Figure \ref{fig:field_fig}, PFML draws embeddings towards other nearby embeddings from the same class rather than distant class embeddings that might be too different to be considered a variation of it. It also helps PFML become significantly more robust to the effect of incorrectly labeled examples as compared to existing methods. Recent work \cite{liu2022debugging, context_icml_2023} shows that such examples are common in metric learning benchmarks. The decay de-emphasizes the contribution of such mislabeled examples, that are typically located away from correctly labeled samples.  We empirically confirm this in Sec. \ref{sec:noise_exp}, demonstrating that methods not adhering to the decay property (i.e., all previous tuplet-based and proxy-based methods) experience significantly greater performance degradation due to label noise compared to PFML.

\subsubsection{Better alignment of proxies with data distribution}\label{sec:alignment_discussion}
The decaying interaction strength helps PFML better align the distribution of proxies with the underlying sample data distribution (of the complete dataset) that they represent. To analyze this, we compare PFML with the closely related contrastive loss \cite{chopra2005learning} that also models pairwise interactions like PFML (defined in Eq. \ref{eq:attract_field_point} \& \ref{eq:repel_field_point} ), and hence can be analyzed similarly by using a potential field. Specifically, we introduce Contrastive Potential based Metric Learning (CPML) by simply extending the contrastive loss by using multiple proxies per class, and subsequently analyzing the resulting loss function through a potential field framework. In CPML, strength of interaction  is directly proportional to the square of the distance from it ($ \| \textbf{r} - \textbf{r}_{0} \|_{2}^{2}$) like in the original contrastive loss . For fair comparison, we keep all other details (use of $\delta$, proxies) as the same, with its associated potential fields denoted by  $\Psi_{att}^{*}\ , \Psi_{rep}^{*}$ ( formally defined in Supplement Sec 1).   We analyze this theoretically using the following proposition and its corollary: 

\textbf{Proposition 1:}\textit{Let $ Z = \{ \textbf{z}_{1} \hdots \textbf{z}_{n} \}$ be a set of sample embeddings belonging to a class, then there exists a $ 0 < \delta < \dfrac{\min_{i,j} \|z_{i} - z_{j} \|_{2}}{2\left(  1 + \frac{1}{n}  \right) }$ and points $\textbf{z}_{min,i}$ within $\delta$ distance from each embedding $z_{i},\  i \in \{1 \hdots n \}$ such that the attractive potential field $\Psi_{att}$ defined using $Z, \delta$ has a minimum at each $\textbf{z}_{min,i}$. The field $\Psi_{att}^{*}$ defined by them does not achieve a minimum at points within $\delta$ distance from all $z_{i}$.}

Proposition 1 is a consequence of the potential field of an embedding dominating in its vicinity over potential fields of other distant embeddings. Its proof relies on the Decay property, which is not satisfied by $\Psi^{*}$ and can be found in Supp. Sec 1. A consequence of this is:

\textbf{Corollary 1:}\textit{ Let $ Z = \{ \textbf{z}_{1} \hdots \textbf{z}_{n} \}$ be a set of sample embeddings from a class exerting an attraction field on a set of proxies $P = \{\textbf{ p}_{1} \hdots \textbf{p}_{m} \}$. Consider the equilibrium distribution $P_{eq}$ of proxies minimizing the potential energy. If the potential field is defined by $\Psi_{\text{att}}$, then the Wasserstein distance $W_{2}$ between $P_{eq}$ and the subset of data they represent is lower than when the potential field is defined by $\Psi_{\text{att}}^{*}$.}

The proof for Corollary 1 relies on Proposition 1 and can be found in the Supplement (Sec. 2).\\

\textbf{Discussion:} As proved in Corollary 1, CPML leads to proxies having a larger distance from the data distribution they represent (measured by $W_{2}$) than our proposed PFML, making the use of proxies in such a framework less effective. This leads to the proxies forming poorer stand-ins for the sample embeddings they represent, reducing the quality of features learned (verified in Sec. \ref{sec:experiments}). This is caused by CPML not following the decay property (which is its only difference in design wrt PFML). Other proxy-based losses like the Proxy NCA \cite{proxy_nca}, Soft Triplet \cite{soft_triplet}, and Proxy Anchor \cite{proxy_anchor} which perform similarly as CPML (see Tab. \ref{tab:eval_cub_cars}), also share with it the property of force increasing with distance (violating the decay property). Due to this, we expect them to similarly have a poorer alignment of proxies with underlying data distribution as compared to PFML (Applying an analysis similar to the one in Proposition 1 to them is considerably more challenging due to the absence of a potential field structure.). We verify this empirically in Sec. \ref{sec:Empirical_distance} by measuring the average distance between proxies and the closest data points ($W_{2}$ distance) for these methods.
\subsection{Advantages in Learning Fine-grained Features}
In contrast with all previous proxy-based methods, PFML (1) Models sample-sample interactions directly using the potential field, gaining fine-grained supervision, and (2) Makes use of an explicit margin-like $\delta$ for proxies, which has been shown to help zero-shot generalization and fine-grained feature learning in tuplet-based methods\cite{wu_sampling_2018}. Turning off the interaction within the $\delta$ radius also prevents a bias towards learning isotropic distributions as is present in other proxy-based methods \cite{roth2022non}, helping us obtain better intra-class resolution. 
\begin{table*}
 \centering
    \setlength{\tabcolsep}{1.5pt}

    \begin{adjustbox}{width=0.9\textwidth}
    
    \begin{tabular}{| l || c  c  c   || c  c  c   || c c c  |}
    \hline
    {\textbf{Benchmarks $ \rightarrow$}} & \multicolumn{3}{c||}{\textbf{CUB-200-2011}} & \multicolumn{3}{c|}{\textbf{Cars-196}} & \multicolumn{3}{c|}{\textbf{SOP}} \\ \hline \hline
    
    Methods $\downarrow$ (Chronological)  & 
        R@1 & R@2 & R@4 &
        R@1 & R@2 & R@4 & 
        R@1 & R@10 & R@100  \\ \hline   
     \rule{0pt}{10pt}\textbf{ResNet50 (512 dim)} & &&& &&& && \\  \hline
    ESPHN~\cite{xuan_easy} & 
        64.9 & 75.3 & 83.5  & 
        82.7 & 89.3 & 93.0  &
        78.3 & 90.7 & 96.3 \\  
    N.Softmax~\cite{norm_softmax} &  
       61.3 & 73.9 & 83.5  &
       84.2 & 90.4 & 94.4  &
       78.2 & 90.6 & 96.2  \\
       DiVA~\cite{diva_dml} & 
        69.2 & 79.3 & - & 
        87.6 & 92.9 & - &
        79.6 & 91.2 & - \\ 
     Proxy NCA++~\cite{proxy_ncapp} &  
       64.7 & - & - & - &
        85.1 & - & - 
         79.6 & - & - \\
    Proxy Anchor~\cite{proxy_anchor} &  
       69.7 & 80.0 & 87.0 &  
        87.7 & 92.9 & 95.8 & 
         - & - & - \\
    DCML-MDW~\cite{DCML} & 
        68.4 & 77.9 & 86.1 &  
        85.2 & 91.8 & 96.0 &
        79.8 & 90.8 & 95.8 \\
    MS+DAS~\cite{das_eccv22} &  
       69.2 & 79.2 & 87.1 &  
        87.8 & 93.1 & 96.0 & 
        80.6 & 91.8 & 96.7 \\ 
    HIST~\cite{lim2022hypergraph} &  
       71.4 & 81.1 & 88.1 & 
         89.6 & 93.9 & 96.4 & 
          81.4 & 92.0 & 96.7 \\ 
    HIER\cite{kim2023hier} &
        70.1 & 79.4 & 86.9  &  
        88.2 & 93.0 & 95.6 & 
        80.2 & 91.5 & 96.6 \\
    HSE-PA~\cite{hse_pa_iccv23} &
        70.6 & 80.1 & 87.1 & 
        89.6 & 93.8 & 96.0 & 
        80.0 & 91.4 & 96.3 \\ 
    CPML (Sec. \ref{sec:discussion})&
        68.3 & 78.7 & 86.2  &  
        85.2 & 91.5 & 95.2 & 
        79.4 & 90.7 & 96.1 \\ \hline
    \textbf{Potential Field (Ours)} &  
       \textbf{73.4 $\pm$ 0.3} & \textbf{82.4$\pm$ 0.1} & \textbf{88.8$\pm$ 0.1} & 
        \textbf{92.7$\pm$ 0.3 }& \textbf{95.5 $\pm$ 0.1} & \textbf{97.6$\pm$ 0.1} &
         \textbf{82.9 $\pm$ 0.2} & \textbf{92.5 $\pm$ 0.2} & \textbf{96.8 $\pm$ 0.1} \\ 
    \hline \hline

     \rule{0pt}{10pt}\textbf{BN Inception (512 dim) } & &&& &&& && \\  \hline
       HTL \cite{HTL} & 
        57.1 & 68.8 & 78.7 &  
        81.4 & 88.0 & 92.7 & 
        74.8 & 88.3 & 94.8 \\
     MultiSimilarity~\cite{multisimilarity_dml} & 
        65.7 & 77.0 & 86.3 & 
        84.1 & 90.4& 94.0 & 
         78.2 & 90.5 & 96.0 \\  
    SoftTriple \cite{soft_triplet} & 
        65.4 & 76.4& 84.5& 
        84.5 & 90.7& 94.5& 
          78.3 & 90.3 & 95.9 \\
    CircleLoss \cite{circle_dml} &  
       66.7 & 77.4 & 86.2 & 
        83.4 & 89.8 & 94.1 &
        78.3 & 90.5 & 96.1  \\
     DiVA~\cite{diva_dml} & 
        66.8 & 77.7 & - & 
        84.1 & 90.7 & - & 
        78.1 & 90.6 & -  \\ 
    ProxyGML~\cite{proxy_gml} &  
       66.6 & 77.6 & 86.4 & 
        85.5 & 91.8 & 95.3 & 
         78.0 & 90.6 & 96.2 \\
    Proxy Anchor~\cite{proxy_anchor} &  
       68.4 & 79.2 & 86.8 & 
        86.1 & 91.7 & 95.0 & 
        79.1 & 90.8 & 96.2 \\ 
     DRML-PA~\cite{DRML} & 68.7 & 78.6 & 86.3 &  
    86.9 & 92.1 & 95.2 & 
    71.5 & 85.2 & 93.0 \\
    MS+DAS~\cite{das_eccv22} &  
       67.1 & 78.11 & 86.4 &  
        85.7 & 91.6 & 95.3 & 
        78.2 & 90.3 & 96.0 \\
   
    HIST~\cite{lim2022hypergraph} &  
       69.7 & 80.0 & 87.3 & 
        87.4 & 92.5 & 95.4 &
         79.6 & 91.0 & 96.2 \\
    DFML-PA~\cite{dfml} &  
       69.3 & - & - &
       88.4 & - & - & 
       - & - & -  \\ 
    HSE-M~\cite{hse_pa_iccv23} &  
       67.6 & 78.0 & 85.8 & 
       82.0 & 88.9 & 93.3 & 
       - & - & - \\ 
    
       PA+niV~\cite{probab_proxy} &  
      69.5 & 80.0 & - &  
       86.4 & 92.0 & - &
       79.2 & 90.4 & - \\ \hline
    \textbf{ Potential Field (Ours)} &  
       \textbf{71.5$\pm$ 0.3} & \textbf{81.2$\pm$0.2} & \textbf{88.3$\pm$0.2} &  
        \textbf{90.1$\pm$0.2} & \textbf{93.9$\pm$0.1} & \textbf{96.3$\pm$0.1} & 
        \textbf{80.6 $\pm$ 0.3} & \textbf{91.8 $\pm$ 0.1} & \textbf{96.4 $\pm$ 0.1}  \\ 
        \hline \hline

      \rule{0pt}{10pt}\textbf{DINO (384 dim)} & &&& &&& && \\ \hline
   DINO \cite{caron2021emerging}  &
    70.8 &81.1 &88.8 & 
    42.9 &53.9 &64.2 & 
    63.4 &78.1 &88.3  \\
    Hyp \cite{ermolov2022hyperbolic}   &
    80.9 &87.6 &92.4 & 
    89.2 &94.1 &96.7 &
    85.1 &94.4 &97.8 \\
    HIER \cite{kim2023hier}   &
    81.1&  88.2&  93.3& 
    91.3&  95.2&  97.1&
    85.7&   94.6&  97.8 \\ \hline 
   \textbf{ Potential Field (Ours)}  &
   \textbf{ 83.1  $\pm$ 0.3} & \textbf{ 89.3 $\pm$ 0.2} & \textbf{ 94.2 $\pm$ 0.1}  &
    \textbf{94.7 $\pm$ 0.1} &  \textbf{96.5 $\pm$ 0.1}&  \textbf{97.8 $\pm$ 0.1}&
    \textbf{86.5 $\pm$ 0.3} &  \textbf{95.1 $\pm$ 0.3} &  \textbf{98.0 $\pm$ 0.2}  \\
 \hline
 \hline
 \rule{0pt}{10pt}\textbf{ViT (384 dim)} & &&& &&& && \\ \hline
     ViT-S \cite{dosovitskiy2021an} &
    83.1 &90.4 &94.4 & 
    47.8 &60.2 &72.2 & 
    62.1 &77.7 &89.0   \\
    Hyp \cite{ermolov2022hyperbolic} &
    85.6 &91.4 &94.8 & 
    86.5 &92.1 &95.3 & 
    85.9 &94.9 &98.1   \\
     HIER \cite{kim2023hier}   &
     85.7&  91.3&  94.4&
     88.3&  93.2&  96.1& 
     86.1&  95.0&  98.0 \\  \hline
     \textbf{Potential Field (Ours)} &
     \textbf{87.8 $\pm$ 0.2} & \textbf{92.6 $\pm$ 0.2} & \textbf{95.7 $\pm$ 0.1} & 
     \textbf{91.5 $\pm$ 0.3} & \textbf{95.2 $\pm$ 0.2} & \textbf{97.4 $\pm$ 0.1} & 
     \textbf{88.2 $\pm$ 0.1} & \textbf{95.7 $\pm$ 0.1} & \textbf{98.6 $\pm$ 0.1}  \\ \hline
    \end{tabular}

    \end{adjustbox}
    
    \vspace{-2mm}
    \caption{
        Comparison of the Recall@$K$ ($\%$) achieved by our method on the CUB-200-2011, Cars-196 and SOP datasets with state-of-the-art baselines under standard settings.  The table reports the average performance and standard deviations of our method over 5 runs. 
        }
    \label{tab:eval_cub_cars}
\end{table*}
\section{Experiments and Results}\label{sec:experiments}

\subsection{Setup}\label{sec:setup}
\textbf{Datasets:} We empirically compare PFML with current state-of-the-art baselines on three public benchmark datasets \textbf{(1)} The Cars-196 dataset \cite{carsdataset} which includes 16,185 images from 196 car model categories, \textbf{(2)} CUB-200-2011 dataset \cite{cubdataset} consisting of 11,788 images from 200 bird species, and \textbf{(3)} Stanford Online Products (SOP) dataset \cite{sopdataset} containing 120,053 images of 22,634 different products sold online. The training and testing splits for the three datasets follow standard zero-shot retrieval-based evaluation protocol \cite{multisimilarity_dml,proxy_anchor, proxy_gml}  of using the first half of the classes for training and the second half for testing.

\textbf{Backbone:} To enable standardized comparison with a wide variety of methods, we evaluate our method using the Inception with batch normalization (IBN) \cite{googlenet}, ResNet-50 \cite{resnet}, ViT \cite{dosovitskiy2021an} and DINO \cite{caron2021emerging} backbones commonly used by DML methods. ImageNet pre-training is used for initialization, and only the last fully connected layer is changed to fit the dimension of the embedding needed. We follow past work \cite{proxy_anchor, proxy_ncapp, soft_triplet} and $\ell_{2}$ normalize the final output. An input image size of $224 \times 224$ is used for all experiments.

\textbf{Training parameters:} We train the backbone for 200 epochs on each dataset. We use the Adam optimizer with the learning rate chosen as $5e^{-4}$. We scale up the learning rate of our proxies by 100 times following \cite{proxy_anchor, proxy_nca, soft_triplet} for better convergence. We choose values of $\delta$ between $[0.1,0.3]$, $\alpha$ between $\{0,6\}$ via cross-validation. We use $M=15$ for the CUB-200 and Cars-196 datasets, while $M=2$ is used for the SOP dataset. All experiments are performed on a single NVIDIA A4000 GPU.

\textbf{Evaluation Settings:} We follow \cite{proxy_anchor,das_eccv22,multisimilarity_dml, kim2023hier} in using $224 \times 224$ sized center crops from $256 \times 256$ images for evaluation. We measure image retrieval performance using Recall@K, which computes the percentage of samples which have a valid similar neighbor (belonging to same class) among its K nearest neighbors.

\subsection{Image Retrieval Performance}\label{sec:results}
As seen in Table \ref{tab:eval_cub_cars}, our method outperforms all other methods on Image Retrieval on the CUB-200-2011, Cars-196 and SOP datasets as measured by the Recall@K metric. It consistently outperforms other methods irrespective of the backbone used, significantly outperforming the best-performing proxy-based methods (which similar to us, only change the loss function) like ProxyAnchor\cite{proxy_anchor}, ProxyGML\cite{proxy_gml} and SoftTriplet \cite{soft_triplet} in terms of Recall@1 (R@1), by more than $5 \%$ on Cars-196, $3.7 \%$ on CUB-200-2011 and $1.5 \%$ on the SOP dataset. It also outperforms the significantly more complex, current state-of-the-art HIST\cite{lim2022hypergraph} by $3.1\%$, $2\%$ and $1.5\%$ in terms of R@1 when using the Resnet-50 backbone on the Cars-196, CUB-200-2011 and SOP datasets respectively. It also significantly outperforms the best tuplet-based loss: the Multi-Similarity loss by $6\%$, $5.8\%$ and $2.4\%$ in terms of R@1 when using the BN Inception backbone on the Cars-196, CUB-200 and SOP datasets respectively. Additionally, our method outperforms CPML (Sec. \ref{sec:alignment_discussion}, a simple combination of the classical contrastive loss with proxies) by  7.5\%, 5.1\%, and 3.5\% on CUB200, Cars and SOP datasets. Note that our performance gains are larger than current SOTA compared to improvements reported by previous efforts on the same benchmarks, despite our intuitive and simpler method.


We also evaluate our method 1) Using the protocol and metrics proposed in \cite{RealityCheck} and (2) When used to learn smaller 128-dim embeddings for more efficient retrieval. Our method still outperforms all SOTA methods in both cases. Details of these experiments are in Supp. Sec. 3. We also show qualitative retrieval results of our method in Supp. Sec 6.

\subsection{Image Retrieval in Presence of Label Noise}\label{sec:noise_exp}
\begin{table*}
  \centering
  \makebox[\textwidth][c]{
  \begin{minipage}[t]{0.62\linewidth}
    \centering
    \setlength{\tabcolsep}{2pt}

   
    \begin{tabularx}{\textwidth}{|X || X  X   || X  X |}
    \hline
    {} & \multicolumn{2}{c||}{\textbf{CUB-200-2011}} & \multicolumn{2}{c|}{\textbf{Cars-196}} \\ \hline \hline

    Methods   & 
        R@1 & R@2 & 
        R@1 & R@2  \\ \hline   
Triplet\cite{triplet_orig} &  
       55.1 & 68.7 & 
        67.5 & 77.9 \\
    MS \cite{multisimilarity_dml} & 
        58.9 & 71.8 & 
        70.4 & 79.8  \\
     PNCA\cite{proxy_nca} &  
       60.1 & 74.7 & 
        74.3 & 82.4  \\
    PA \cite{proxy_anchor} &  
       60.7 & 75.1 &
        76.9 & 83.1  \\
    HIST\cite{HIST_code} &  
       59.7 & 74.6 &
         72.9 & 81.8   \\ 
     \hline
 
    \textbf{Ours} &  
       \textbf{66.7 $\pm$ 0.6} & \textbf{76.9$\pm$ 0.3} & \textbf{84.5$\pm$ 0.5} & \textbf{88.6$\pm$ 0.3}  \\ 
    \hline 
    \end{tabularx}
   
    \caption{A comparison of Recall@$K$ ($\%$) achieved by our method with SOTA baselines on CUB-200-2011 and Cars-196 datasets  with label noise when using a Resnet-50 backbone (512-dim). Performance and std. deviations reported are over 5 runs.
        }
    \label{tab:eval_noise}

    \end{minipage}

    \hspace{0.02\textwidth}
    
    \begin{minipage}[t]{0.36\linewidth}
    \centering
    \begin{tabular}{|c||c|}
    \hline
    \textbf{Method} & \textbf{Average} $\textbf{W}_{2}$ \\ \hline
        S.Triplet \cite{soft_triplet} & $0.24 \pm 0.04$ \\
        PNCA \cite{proxy_anchor}& $0.36 \pm 0.02$\\ 
        PA \cite{proxy_anchor}& $0.38 \pm 0.03$\\ 
        CPML (Sec. \ref{sec:discussion})  & $0.26 \pm 0.03$ \\ \hline \hline
       \textbf{ Ours} &  $\textbf{0.16} \pm \textbf{0.03}$\\ \hline
    \end{tabular}
    
    \caption{Average Wasserstein distance  between proxies and the subset of data closest to them for different methods on CUB-200-2011 dataset. Mean and std. deviations are over 5 runs.}
    
    \label{tab:w2_dist}
    \end{minipage}
    }
    \vspace{-10pt}
\end{table*}
    
\begin{figure*}
    \centering
    \includegraphics[width = 0.97\textwidth]{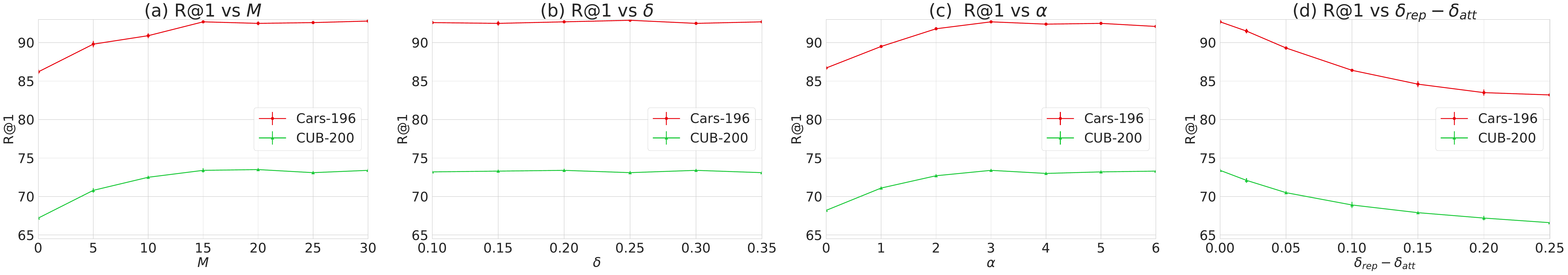}
    \caption{Variation in Recall@1 with $M$, $\delta$, $\alpha$ and $\delta_{rep} - \delta_{att}$ on the Cars-196 and CUB-200-2011 datasets. Error bars represent std. deviations over 5 runs.}
    \label{fig:ablation_all}
    \vspace{-20pt}
\end{figure*}
We evaluate the robustness of our method vs. state-of-the-art baselines in the realistic scenario of label noise (e.g., due to mislabeling, which is very common, particularly in fine-grained datasets such as those used for DML). Specifically, we randomly corrupt labels of 20\% of samples in the training set. As seen in the results in Table \ref{tab:eval_noise}, this leads to a significant decline in the performance of all methods. But our method is the least affected by the noise.
Other methods like the triplet \cite{harwood_smart_2017} or ProxyAnchor\cite{proxy_anchor} suffer much greater declines because they are affected more by distant points represented by the mislabeled examples. Our method significantly outperforms Proxy Anchor \cite{proxy_anchor} (the second-best method) by $6\%$ and $7.6\%$ in terms of R@1 on the CUB-200-2011 and Cars-196 datasets. 
\subsection{Distance Between Proxies and Sample Embeddings}\label{sec:Empirical_distance}
 We measure the average (over a complete epoch) Wasserstein distance $W_{2}$ between proxies and the subset of training data represented by/closest to them \footnote{For methods using a single proxy per class, this equals the distance between a proxy and the closest embedding belonging to its class} for different proxy-based methods. We train networks on the CUB-200-2011 using the settings in Section \ref{sec:setup} and use the \textit{same number of proxies} for all methods utilizing multiple proxies. As seen in Table \ref{tab:w2_dist}, our method shows 33.3\% lower $W_{2}$ distance between the proxies and the subset of data than the second best SoftTriplet \cite{soft_triplet} loss, empirically validating our discussion in Sec. \ref{sec:alignment_discussion}.
\subsection{Ablation studies}\label{sec:ablation}
 In all experiments, we use the parameters mentioned in Section \ref{sec:setup} with a ResNet-50 backbone on both the Cars-196 and CUB-200-2011 datasets.

\textbf{Effect of proxies and their number:}
Proxies augment the potential field, and the number of proxies per class $M$ decides the granularity at which the proxies model the distribution of sample embeddings. We study their effect on performance by varying $M$ between $[0,30]$. From the results plotted in Figure \ref{fig:ablation_all}(a) we observe that (1) Using proxies to augment the field boosts performance (vs using no-proxies) and (2) Network performance remains stable for a wide range of M, decreasing slightly for small values of M (still outperforming most other proxy-based methods). This behavior is expected as using potential field created using a very small M would not be able to effectively model underlying sample distribution, leading to worse supervision. We also observe that using a larger number of proxies per class does not significantly affect training times, as bulk of computation is dominated by the network. For further details, please refer to Supp. Sec. 5. 

\textbf{Effect of varying }$\delta$ :
We conduct an empirical study to determine the effect of varying the radius parameter $\delta$ on performance by varying it over $[0.1, 0.35]$. As seen in Figure \ref{fig:ablation_all}(b), we observe that performance remains relatively stable for a wide range of $\delta$ values, demonstrating our method's robustness to it.

\textbf{Effect of potential decay} $\alpha$:
The decay parameter $\alpha$ determines the rate at which the potential field of an embedding decays with distance. We study its effect by varying it over $\left[0,6\right]$. As seen in Figure \ref{fig:ablation_all}(c), performance is worse when the potential field does not decay ($\alpha = 0$). A field that decays strongly $\alpha \in [3,6]$ seems to perform better than one with a much milder decay ($\alpha \in \left[1,2\right]$). The likely reason for this trend is that when the potential field has no decay /decays mildly, the force on an embedding would still be dominated by a large number of distant sample embeddings rather than those in its vicinity.

\textbf{Restricting repulsion to a distance of} $\delta$:
Unlike most proxy-based methods, our design of potential field $\Psi_{rep}$ limits the range of repulsion to within $\delta$ distance from embeddings or proxies. We examine extending this range of repulsion beyond the $\delta$ radius by having $ \delta_{rep} > \delta_{att}$  where $\delta_{att}$ and $\delta_{rep}$ represent the values of $\delta$ used for attraction (Eq.\ref{eq:attract_field_point}) and repulsion (Eq.\ref{eq:repel_field_point}) fields. Fixing $\delta_{att} = 0.2$, we increase $\delta_{rep}$ to study the effect. As shown in Figure \ref{fig:ablation_all}(d), his reduces performance, validating our design. This is due to individual potentials repelling dissimilar points to a distance of at least $\delta_{rep}$, greater than the distance $\delta_{att}$ within which all points of the same class are attracted to. Using $ \delta_{rep} > \delta_{att}$ enforces excessive separation between dissimilar examples, compressing intra-class features more than needed.
\vspace{-5pt}
\section{Limitations}
\vspace{-5pt}
PFML relies on proxies to augment potential fields in each batch, hence sharing the main limitation of proxy-based methods: with a large number of classes, the computational cost during training increases due to linear scaling of proxies with class count.
Despite this, proxy-based methods have been widely used in applications with a large number of classes like face recognition (e.g. \cite{arcface} on 500,000 classes), likely because (1) proxies only affect training, not inference complexity and (2) The parameters/compute complexity added by the proxies is relatively low for the performance gains they offer, especially when such compute complexity is compared to the forward and backward passes of the neural network that consume bulk of compute in DML.
\vspace{-5pt}
\section{Conclusion}
\vspace{-5pt}
PFML introduces a novel paradigm to capture interactions between all samples through a continuous potential field, where strength of sample influence decreases with distance.
 This reversal of the common model of increasing influence with distance improves robustness to label noise and better aligns proxy distributions with sample embeddings.  The overall design of our method is validated by its performance on standard image retrieval benchmarks, where it outperforms state-of-the-art methods both in the standard noise-free setting and in the presence of label noise unavoidable in real-world datasets. 

\section{Acknowledgement}\vspace{-5pt}
The support of the Office of Naval Research under grant N00014-20-1-2444 and of USDA National Institute of Food and Agriculture under grant 2020-67021-32799/1024178 are gratefully acknowledged. This research used the Delta system at NCSA through allocation CIS230360 from the ACCESS program which is which is supported by National Science Foundation grants \#2138259, \#2138286, \#2138307, \#2137603, and \#2138296.

{
    \small
    \bibliographystyle{ieeenat_fullname}
    \bibliography{egbib}
}
 \onecolumn
 \setcounter{section}{0} 
\begin{center}
    {\Large \textbf{Supplementary Material: Potential Field Based Deep Metric Learning}}
\end{center}

In this supplementary material, we provide additional information which we could not fit in the space available in the main paper. We do so in six sections: Sections \ref{sec:sup_proof_proposition1} \& \ref{sec:sup_proof_corollary1} contain proofs for Proposition 1 and Corollary 1 respectively; Section \ref{sec:supp_experiments} contains additional empirical studies that further validate the effectiveness of our method ( Sec. \ref{sec:small_embed}: using a smaller embedding size, Sec. \ref{sec:results_mlrc}: Results when using the MLRC protocol \cite{RealityCheck}); Section \ref{sec:hyperparam} provides implementation details about the hyperparameters used; Section \ref{sec:compute} empirically compares the time complexity of our method with other methods; Section \ref{sec:visual} provides qualitative retrieval results and a t-sne visualization of the embedding space learned by PFML. 

\section{Proof of Proposition 1}\label{sec:sup_proof_proposition1}
\textbf{Proposition 1:}\textit{Let $ Z = \{ \textbf{z}_{1} \hdots \textbf{z}_{n} \}$ be a set of sample embeddings belonging to a class, then there exists a $\displaystyle 0 < \delta < \dfrac{\min_{i,j} \|z_{i} - z_{j} \|_{2}}{2\left(  1 + \frac{1}{n}  \right) }$ and a distance $z_{min,i} \leq \delta$  for each embedding $z_{i},\  i \in \{1 \hdots n \}$ such that the attractive potential field $\Psi_{att}$ ( Eq. 4 from paper ) defined using $(Z, \ \delta)$ when restricted to a radial line  from $z_{i}$ has a minimum at each $z_{min,i}$. The field $\Psi_{att}^{*}$  (by CPML, the interaction strength increasing potential defined in Eq. 2) defined by $Z$ does not achieve such a minimum at points within $\delta$ distance from all $z_{i}$.}
\\
\textbf{Proof:} We first define the potential field at point $\textbf{z}$ caused by an individual embedding  $ \textbf{z}_{i}$, $\psi_{att}(\textbf{z}, \textbf{z}_{i})$ and $\psi_{att}^{*}(\textbf{z}, \textbf{z}_{i})$ are defined as:
\begin{align}
     \psi_{att}(\textbf{z}, \textbf{z}_{i}) := \begin{dcases} 
    -\dfrac{1}{\delta^{\alpha}} & \textrm{if } \| \textbf{z} - \textbf{z}_{i} \|_{2} < \delta \\
    -\dfrac{1}{\| \textbf{z} - \textbf{z}_{i}  \|_{2}^{\alpha}} & \textrm{otherwise.} 
    \end{dcases} \label{eq:pair_field_def}
 \end{align}
\begin{align}
     \psi_{att}^{*}(\textbf{z}, \textbf{z}_{i}) := \begin{dcases} 
   \delta^{2} & \textrm{if } \| \textbf{z} - \textbf{z}_{i} \|_{2} < \delta \\
   \| \textbf{z} - \textbf{z}_{i}  \|_{2}^{2} & \textrm{otherwise.} 
    \end{dcases} \label{eq:pair_field_def2}
 \end{align}
The potential fields created by all data points are:
\begin{align}
    \Psi_{att}(\textbf{z}) = \sum_{i=1}^{n}  \psi_{att} (\textbf{z}, \textbf{z}_{i}) \label{eq:our_field_def} \\ 
    \Psi_{att}^{*}(\textbf{z}) = \sum_{i=1}^{n}  \psi_{att}^{*} (\textbf{z} , \textbf{z}_{i}) \label{eq:other_field_def}
\end{align}
We prove the proposition in two parts, first proving the assertion for $\Psi_{att}(\textbf{z})$ in Part 1, and then moving on to proving the assertion for $ \Psi_{att}^{*}(\textbf{z})$  in Part 2.
\subsection{Part 1: Proof for $\Psi_{att}(\textbf{z})$}
To prove that  $\Psi_{att}(\textbf{z})$ achieves a minimum in the radial direction at a distance $z_{min,i} \leq  \delta$  distance of each embedding $z_{i}$, we observe that $\Psi_{att}$ is continuous and bounded within the $\delta $ hyper-spheres $S_{i} = \|\textbf{z} - \textbf{z}_{i} \| \leq \delta$. Each hypersphere is a closed bounded set. 

This enables us to apply the Extreme Value Theorem (EVT) \cite{extreme_value_theorem} to $\Psi_{att}$ on $S_{i}$. Using EVT, we note that $\Psi_{att}$ achieves a minimum $\Psi_{att}(\textbf{z}^{*})$ on the set $S_{i}$ at some distance $z_{min,i}$. 

$z_{min,i}$ may either be $< \delta$  (minimum inside the sphere) or $= \delta$ (minimum on the boundary). We analyze both cases separately, proving that the minimum for $\Psi_{att}$,  $z_{min,i}$ on $S_{i}$ is also a minimum for $\Psi_{att}$ on the embedding space $ R^{D}$ (D = embedding dimension):\\
\\
\textbf{Case 1:}  If minimum $\textbf{z}^{*}$ lies inside the sphere $S_{i}$, then $z_{min,i}$ is a minimum for $\Psi_{att}$ on $R^{D}$ too because $S_{i} \subset R^{D}$ (D = embedding dimension).\\
\textbf{Case 2:} If $z_{min,i} = \delta$  (minimum $\textbf{z}^{*}$  lies on the border of sphere $S_{i}$). The proof for this intuitively relies on the fact that the derivative of the potential field of $\textbf{z}_i$ ($\psi_{\text{att}}(\textbf{z}, \textbf{z}_i)$) outside $S_{i}$, around $\textbf{z}^*$ is positive, and the derivative (i.e., interaction strength or force) increases as we choose a smaller $\delta$. Consequently, the derivative of the potential field of $\textbf{z}_i$ dominates the force applied at $\textbf{z}^*$ compared to the force applied by potentials of other embeddings (which would have decayed). Hence, with a small enough delta, it is possible to ensure that the potential at points outside $S_{i}$ is larger than the one on the boundary. 

To formally handle this case, we first define $\Psi_{att}|_{R_{i}}$ as the restriction of $\Psi_{att}$ to the line $R_{i} = \{ \mathbf{z_{i}} + t\  \mathbf{z}^{*} \mid t \in \mathbb{R} \}.$ Then by the definition of a local minimum

$\textbf{z}^{*}$ is a minimum of $\Psi_{att}|_{R_{i}}$  iff there exists an $\epsilon > 0$ such that $\Psi_{att}|_{R_{i}}(\textbf{z}^{*}) \leq \Psi_{att}|_{R_{i}}(\textbf{z}^{'})$ for all $\textbf{z}^{'}\in R_{i}$. 
We note that for any $\textbf{z}^{'}\in S_{i}$,  $\Psi_{att}|_{R_{i}}(\textbf{z}^{*}) \leq \Psi_{att}|_{R_{i}}(\textbf{z}^{'})$ is trivially true by the definition of $\textbf{z}^{*}$. For $\textbf{z}^{'}\notin S_{i}$, we use the Taylor expansion of  $\Psi_{att}|_{R_{i}}$  to find such an $\epsilon$ and prove that such a $\textbf{z}^{*}$ is also a minimum for $\Psi_{att}|_{R_{i}}$ in $R^{D}$  .

 Specifically, the Taylor expansion for $\Psi_{att}$ in the radial direction (co-ordinates centered at $z_{i}$) is given as: 

\begin{align}
  \Psi_{att}|_{R_{i}}(\textbf{z}^{'}) & = \Psi_{att}|_{R_{i}}(\textbf{z}^{*}) + ( \ \mathbf{z^{'}} -  \mathbf{z}^{*}). \left(   \left.  \frac{\partial  \Psi_{att}|_{R_{i}}}{\partial  \mathbf{z^{'}} } \right|_{ \mathbf{z}^{*}}     \hat{\mathbf{r}} \right) + \text{err}(\mathbf{z^{'}} -  \mathbf{z}^{*})  \label{eq:prop1_part1_taylor} \\
   \text{here  } \hat{\mathbf{r}} & \text{ is the unit radial vector pointing toward }  \textbf{z}^{*}\text{ centered at } \textbf{z}_{ i}
\end{align}

Expanding the partial derivative, we get:
      \begin{align}
       \frac{\partial  \Psi_{att}|_{R_{i}}}{\partial  \mathbf{z^{'}} }  & =  \sum_{i=1}^{n} \frac{\partial \psi_{att} (\mathbf{z^{'}}, \textbf{z}_{i})|_{R_{i}} } {\partial\mathbf{z^{'}}}  
     \end{align}
Evaluating it at $\textbf{z}^{*}$, we get:

\begin{align}
     \frac{\partial  \Psi_{att}|_{R_{i}} }{\partial  \mathbf{z^{'}} }  \left( 
 \mathbf{z}^{*}  \right) & =  
 \dfrac{\alpha}{  \| \textbf{z}^{*}  - \textbf{z}_{i}  \|_{2}^{\alpha+1}  } + \gamma_{i}\nonumber \end{align}
    \begin{align}
       \frac{\partial  \Psi_{att}|_{R_{i}} }{ \partial  \mathbf{z^{'}} }  \left(\mathbf{z}^{*}\right) & =  \dfrac{\alpha}{\delta^{\alpha+1}} \hat{\mathbf{r}} + \gamma_{i} \nonumber \\
     \text{where } \gamma_{i} & = \sum_{j=1 , \neq i}^{n} \frac{\partial \psi_{att} (\mathbf{z^{'}}, \textbf{z}_{i})|_{R_{i}} } {\partial\mathbf{z^{'}}}  \nonumber
\end{align}
Now $\dfrac{\alpha}{\delta^{\alpha+1}} > \gamma_{i}$ for all $\delta < \left( \dfrac{\alpha}{\gamma_{i}} \right)^{\frac{1}{\alpha +1}}$. So, 
\begin{align}
   ( \ \mathbf{z^{'}} -  \mathbf{z}^{*}). \left(   \left.  \frac{\partial  \Psi_{att}|_{R_{i}}}{\partial  \mathbf{z^{'}} } \right|_{ \mathbf{z}^{*}}     \hat{\mathbf{r}} \right) = \|\mathbf{z^{'}} -  \mathbf{z}^{*}\| \left(  \dfrac{\alpha}{\delta^{\alpha+1}} + \gamma_{i} \right) > 0 \ \ \  
    \forall \delta < \left( \dfrac{\alpha}{\gamma_{i}} \right)^{\frac{1}{\alpha +1}}  \label{eq:taylor_inter1}
\end{align}

Also, as $\epsilon \to 0$,  $ \dfrac{\text{err}(\mathbf{z^{'}} -  \mathbf{z}^{*})}{   ( \ \mathbf{z^{'}} -  \mathbf{z}^{*}). \left(   \left.  \frac{\partial  \Psi_{att}|_{R_{i}}}{\partial  \mathbf{z^{'}} } \right|_{ \mathbf{z}^{*}}     \hat{\mathbf{r}} \right) } \to 0 $ as those are higher order terms. \\ \\
Simplifying the Taylor expansion (Eq. \ref{eq:prop1_part1_taylor})
\begin{align}
 \Psi_{att}|_{R_{i}}(\textbf{z}^{'}) - \Psi_{att}|_{R_{i}}(\textbf{z}^{*}) & =  + ( \ \mathbf{z^{'}} -  \mathbf{z}^{*}). \left(   \left.  \frac{\partial  \Psi_{att}|_{R_{i}}}{\partial  \mathbf{z^{'}} } \right|_{ \mathbf{z}^{*}}     \hat{\mathbf{r}} \right) + \text{err}(\mathbf{z^{'}} -  \mathbf{z}^{*}) \nonumber  \\
  &=  ( \ \mathbf{z^{'}} -  \mathbf{z}^{*}). \left(   \left.  \frac{\partial  \Psi_{att}|_{R_{i}}}{\partial  \mathbf{z^{'}} } \right|_{ \mathbf{z}^{*}}     \hat{\mathbf{r}} \right) \text{ as } \epsilon \to 0   \nonumber \end{align}
\begin{align}
   \therefore \Psi_{att}|_{R_{i}}(\textbf{z}^{'}) - \Psi_{att}|_{R_{i}}(\textbf{z}^{*})& > 0 \text{ (Using Eq. \ref{eq:taylor_inter1})}
\end{align}
So for $\textbf{z}^{'} \notin S_{i}$, $\Psi_{att}(\textbf{z}_{min,i}) \leq \Psi_{att}(\textbf{z}^{'})$ for all $\|\textbf{z}^{'} - \textbf{z}^{*} \| < \epsilon$.\\ \\
Therefore, by definition $\textbf{z}^{*}$ is a minimum of $\Psi_{att}|_{R_{i}}$ on $R^{D}$ in Case 2 too. \\
Hence,  $\Psi_{att}$ when restricted to a radial line  from $z_{i}$ has a minimum at a distance $z_{min,i} \leq \delta$ from $\textbf{z}_{i} \forall i \in \{1 \hdots n\}$ for $\delta < argmin_i \left( \dfrac{\alpha}{\gamma_{i}} \right)^{\frac{1}{\alpha +1}} $. $z_{min,i}$ lies within $\delta$ distance of the embeddings $z_{i}$ by definition of the hyperspheres $S_{i}$. 

\subsection{Part 2: Proof for $\Psi_{att}^{*}(\textbf{z})$}
Outside the hyperspheres $S_{i} = \|\textbf{z} - \textbf{z}_{i} \| \leq \delta$, $ \nabla\Psi_{att}^{*}(\textbf{z})$ ($\textbf{z} \notin S_{i}$)is given by:
\begin{align}
   \nabla\Psi_{att}^{*}(\textbf{z}) &=  \sum_{i=1}^{n}  \nabla \psi_{att}^{*} (\textbf{z} - \textbf{z}_{i}) \nonumber \\
   \nabla\Psi_{att}^{*}(\textbf{z}) &=  \sum_{i=1}^{n}   2(\textbf{z} - \textbf{z}_{i}) = 2n\textbf{r} - \sum_{i=1}^{n}  2\textbf{z}_{i} = 2n \left(\textbf{z} - \sum_{i=1}^{n}  \dfrac{\textbf{z}_{i}}{n} \right) \label{eq:grad_def}
\end{align}
It achieves a single minimum at $\textbf{z}_{global min} = \dfrac{\sum_{i=1}^{n}  \textbf{z}_{i}}{n} $.\\
Now define differentiable extensions of the potentials- 
\begin{align}
     \psi_{att}^{**}(\textbf{z} - \textbf{r}_{0}) = 
   \| \textbf{z} - \textbf{r}_{0}  \|_{2}^{2} \label{eq:diff_poten}
 \end{align}
To calculate $ \nabla\Psi_{att}^{*}(\textbf{z})$ inside the hyperspheres ($\textbf{r} \in S_{i}$), we note that   no two hyper spheres $S_{i}$ intersect with each other as $\delta < 0.5\min_{i,j} \|z_{i} - z_{j} \|_{2}$. Observing that for $\textbf{r} \in S_{i}$
\begin{align}
       \nabla\Psi_{att}^{*}(\textbf{z}) =  \sum_{j=1 ,j \neq i}^{n}  \nabla \psi_{att}^{**} (\textbf{z} - \textbf{z}_{j})     \nonumber
\end{align}
Using triangle inequality for $\textbf{z} \in S_{i}$:
\begin{align}
      \| \sum_{j=1 }^{n}  \nabla \psi_{att}^{**} & (\textbf{z} - \textbf{z}_{j}) \| - \|  \nabla \psi_{att}^{**} (\textbf{z} - \textbf{z}_{i}) \| \leq \| \sum_{j=1 ,j \neq i}^{n}  \nabla \psi_{att}^{*} (\textbf{z} - \textbf{z}_{j}) \| = \| \nabla\Psi_{att}^{*}(\textbf{z}) \|   \nonumber \\
      \text{Using } &  \|  \nabla \psi_{att}^{**} (\textbf{z} - \textbf{z}_{i}) \| \leq 2 \delta \text{ (from Eq. \ref{eq:diff_poten})}  \nonumber \\
      \| \nabla\Psi_{att}^{*}(\textbf{z}) \|  &\geq  \| \sum_{j=1}^{n}  \nabla \psi_{att}^{**} (\textbf{z} - \textbf{z}_{j}) \| -2 \delta \nonumber
\end{align}
We know that the RHS term $> 0$  because $ \| \sum_{j=1}^{n}  \nabla \psi_{att}^{**} (\textbf{z} - \textbf{z}_{j}) \| > 2 \delta$ for 
\begin{align}
    \| \textbf{z} - \sum_{i=1}^{n}  \dfrac{\textbf{z}_{i}}{n} \| > \dfrac{\delta}{n} \label{eq:max_radius}
\end{align} using Equation \ref{eq:grad_def}. For all such $\textbf{z}$:
\begin{align}
    \| \nabla\Psi_{att}^{*}(\textbf{z}) \|  &> 0 \label{eq:grad_non_zero}
\end{align}
At most one sphere $S_{i}$ has an $\textbf{r}_{i} \in S_{i}$ not satisfying Equation \ref{eq:max_radius}. We prove this by contradiction. Assume that another sphere $S_{j}$ has $\textbf{r}_{j} \in S_{j}$ not satisfying Equation \ref{eq:max_radius}. Now the distance between $\textbf{r}_{i}$ and $\textbf{r}_{j}$ satisfies:
\begin{align}
    \| \textbf{r}_{i}- \textbf{r}_{j} \| \geq d_{min} - 2\delta \label{eq:sphere_dupli}
\end{align}
here $\displaystyle d_{min} = \min_{i,j} \|z_{i} - z_{j} \|_{2} $
Using the fact that $\displaystyle \delta < \dfrac{d_{min}}{2\left(  1 + \frac{1}{n}  \right) }$ by definition and substituting for $d_{min}$ in equation \ref{eq:sphere_dupli} we get:
\begin{align}
 \| \textbf{r}_{i}- \textbf{r}_{j} \| > 2\delta \left(  1 + \dfrac{1}{n}  \right)  - 2\delta    \nonumber \\
  \| \textbf{r}_{i}- \textbf{r}_{j} \| > \dfrac{2 \delta}{n} \nonumber
\end{align}
Both $\textbf{
}_{i}$ and $\textbf{r}_{j}$ cannot satisfy equation \ref{eq:max_radius} as all points satisfying it lie within a sphere of radius $\frac{\delta}{n}$, and distance between $\textbf{r}_{i}, \textbf{r}_{j}$ is more than the maximum distance between points in a sphere, that is $2\frac{\delta}{n}$. Hence, no such $\textbf{r}_{j} \in S_{j}$ can exist.

Hence for any other hypersphere $S_{j}, i \neq j$,  for all $\textbf{z}$ we have $  \| \nabla\Psi_{att}^{*}(\textbf{z}) \|  > 0 $, and hence no minimum exists within them. Hence, proved.

\section{Proof of Corollary 1}\label{sec:sup_proof_corollary1}
\textbf{Corollary 1:}\textit{ Let $ Z = \{ \textbf{z}_{1} \hdots \textbf{z}_{n} \}$ be a set of sample embeddings belonging to a class exerting an attraction field on a set of proxies $P = \{\textbf{ p}_{1} \hdots \textbf{p}_{m} \}$. Consider the equilibrium distribution $P_{eq}$ of proxies minimizing the potential energy. If the potential field is defined by $\Psi_{\text{att}}$, then the Wasserstein distance $W_{2}$ between $P_{eq}$ and the subset of data they represent is lower than when the potential field is defined by $\Psi_{\text{att}}^{*}$.} 
\\
\textbf{Proof:}
Let the potential fields $\Psi_{\text{att}}$ and $\Psi_{\text{att}}^{*}$ be given by Equations \ref{eq:our_field_def} and \ref{eq:other_field_def} respectively.
\begin{align}
    \psi_{total}(\textbf{r}) = \sum_{i=1}^{n}  \psi (\textbf{r} - \textbf{z}_{i})
\end{align} 
where $\psi$ is defined using Eq 4 from the paper(att and class subscript j omitted for clarity).
The potential energies of the proxy distribution $P_{eq}$ in these potential fields, $ \mathcal{U}_{proxy}$ and $ \mathcal{U}_{proxy}^{*}$ respectively are given by :
\begin{align}
    \mathcal{U}_{proxy} = \sum_{p_{i} \in P_{eq}} \Psi_{\text{att}}(\textbf{p}_{i}) \nonumber \\
    \mathcal{U}_{proxy}^{*} = \sum_{p_{i} \in P_{eq}} \Psi_{\text{att}}^{*}(\textbf{p}_{i}) \nonumber
\end{align}
At equilibrium, each proxy migrates to the nearest minimum in the field. The subset of data the proxies represent are given by the subset of m data points $Z_{subset} = z_{f(k)} , \ k \in \{1 \hdots m\}; Z_{subset} \subset Z$ which is the closest in Wasserstein distance $W_{2}$ from the proxies.

\textbf{Case 1:} Let the field be defined  by $\Psi_{\text{att}}$.  \textbf{Assumptions:} As the distances of proxies $\textbf{p}_{k}, k \in {1 \hdots m}$ from embeddings $z_{i}$ are minimized, we assume that they migrate to a distance of $d_{min,f(k)}$ from the nearest embedding in the potential field denoted by $\textbf{z}_{min,f(k)}$ which is located within $\delta $ distance of data point $z_{f(k)}$ ( using proposition 1 ). Assuming that the proxies are initialized using a normal distribution (commonly used) and $m << n$, which is typically true, we ignore the probability of more than one proxy going to the same minimum $z_{min,f(k)}$ (so $f(k)$ is one-one). Therefore:
\begin{align}
    W_{2}(P_{eq}, Z_{subset}) & = \inf_{\pi}\left( \dfrac{1}{m} \sum_{k=1}^{m} \| p_{k} - \textbf{z}_{\pi(k)} \|_{2} \right) \nonumber \\ 
    \text{ here the } & \text{infimum is over all permutations $\pi$ of k elements} \nonumber \\
    W_{2}(P_{eq}, Z_{subset}) & = \left( \dfrac{1}{m} \sum_{k=1}^{m} \| \textbf{z}_{min,f(k)} - \textbf{z}_{f(k)} \|_{2} \right) \nonumber \\
    \text{Using proposition 1 }  \nonumber \\
    W_{2}(P_{eq}, Z_{subset}) & \leq  \left( \dfrac{1}{m} \times (m \delta)  \right)  \nonumber \\
    W_{2}(P_{eq}, Z_{subset}) & \leq \delta
\end{align}
Hence when the field is defined by $\Psi_{\text{att}}$ we have $W_{2}(P_{eq}, Z_{subset}) \leq \delta$.\\
\textbf{Case 2: } Let the field be defined  by $\Psi_{\text{att}}^{*}$. The proxies $\textbf{p}_{k}, k \in {1 \hdots m}$ migrate to the nearest minimum in the potential field denoted by $\textbf{z}_{min,f(k)}$. Using Proposition 1 proved before, we know that all minima satisfy $\displaystyle \min_{j} \| \textbf{z}_{min,f(k)} -\textbf{z}_{j}\| > \delta$ for all $j$ except at most one $j=j'$ for which let $k =k'$.\\

First, we prove the corollary for the case if there exists such a $j=j'$. Let $\textbf{z}_{g(k)} , k = \{1 \hdots m \}$ represent the ordered subset of data embeddings that minimize the $W_{2}$ distance metric with the proxies $\textbf{p}_{k} \in P_{eq}$. From proposition 1, we know that:
\begin{align}
    \| \textbf{z}_{min,f(k)} -\textbf{z}_{j^{'}}\|_{2} \leq \delta  \nonumber \\
\end{align}
we also have: 
\begin{align}
\| \textbf{z}_{g(k)} - \textbf{z}_{j^{'}} \|_{2} \geq \min_{i,j} \|\textbf{z}_{i} - \textbf{z}_{j} \|_{2}  \geq 2\delta \left(1 + \dfrac{1}{m} \right) \nonumber \\ 
\end{align}
Using the triangle inequality on the above 2 inequalities and substituting, we get:
\begin{align}
\| \textbf{z}_{min,f(k)} - \textbf{z}_{g(k)}\|_{2} & \geq \| \textbf{z}_{g(k)} -\textbf{ z}_{j^{'}} \|_{2} -  \| \textbf{z}_{min,f(k)} -\textbf{z}_{j^{'}}\|_{2} \nonumber \\
\| \textbf{z}_{min,f(k)} - \textbf{z}_{g(k)}\|_{2} & \geq 2\delta \left(1 + \dfrac{1}{m} \right) -  \delta \nonumber \\
\| \textbf{z}_{min,f(k)} - \textbf{z}_{g(k)}\|_{2} & \geq \delta + \dfrac{2\delta}{m}   \label{eq:main_ineq2}
\end{align}
    
The $W_{2}^{*}$ distance is given by:
\begin{align}
   &  W_{2}^{*}(P_{eq}, Z_{subset}) = \inf_{\pi}\left( \dfrac{1}{m} \sum_{k=1}^{m} \| \textbf{p}_{k} - \textbf{z}_{\pi(k)} \|_{2} \right) \nonumber \\ 
    & \text{ here the infimum is over all permutations $\pi$ of k elements} \nonumber \\
    & W_{2}^{*}(P_{eq}, Z_{subset}) = \left( \dfrac{1}{m} \sum_{k=1}^{m} \| \textbf{z}_{min,f(k)} - \textbf{z}_{g(k)} \|_{2} \right) \nonumber \\
   & W_{2}^{*}(P_{eq}, Z_{subset})  = \left( \dfrac{1}{m} \sum_{k=1, k \neq k^{'} }^{m} \| \textbf{z}_{min,f(k)} - \textbf{z}_{g(k)} \|_{2} \right) + \dfrac{1}{m} \| \textbf{z}_{min,k^{'}} - \textbf{z}_{j'} \|_{2} \nonumber \\
    & \text{Using equation \ref{eq:main_ineq2} and propositon 1 }  \nonumber \\
   & W_{2}^{*}(P_{eq}, Z_{subset}) \geq  \dfrac{1}{m} \times \left( (m-1)\delta  + \dfrac{2(m-1)\delta}{m} \right)  \nonumber \\
   & W_{2}^{*}(P_{eq}, Z_{subset}) > \delta
\end{align}

Hence the $W_{2}$ distance between $P_{eq}$ and $Z_{subset}$ when the field is given by $\Psi_{att}$ is $W_{2}(P_{eq}, Z_{subset}) \leq \delta$ while their distance $W_{2}^{*}(P_{eq}, Z_{subset})$ when the field is $> \delta$ as proved above. So $W_{2}^{*}(P_{eq}, Z_{subset}) > W_{2}(P_{eq}, Z_{subset})  $. Hence, proved.


\section{Additional Experimental Results}\label{sec:supp_experiments}
In this section of the supplement, we provide additional experiments that we could not fit in the space available in the main paper. These empirical studies further validate the effectiveness of our method.
\subsection{Performance using Small Embedding size}\label{sec:small_embed}
\textbf{Context:} In Section 4.2, we presented results on image retrieval using an embedding space of dimension 512.
However in certain settings, learning embeddings in a lower dimension space might be more useful, such as in settings where limited storage is available for storing image embeddings. While this lowers the image retrieval performance, as is to be expected, it allows for a trade-off between available memory/compute resources and the accuracy of retrieval. Hence, we compare the performance of our method in learning a lower dimensional embedding space with recent state-of-the-art baselines. 

\textbf{Experiment:} We train a ResNet-50 network with its embedding size set to 64 (a commonly used setting) on the Cars-196 \cite{carsdataset}, CUB-200-2011\cite{cubdataset} and SOP \cite{sopdataset} datasets. 

\textbf{Results:} As seen in Tables \ref{tab:eval_small_embed} and \ref{tab:eval_small_sop}, we observe that our method is able to outperform all other methods at this task; specifically, it outperforms strong Proxy-based baselines like ProxyAnchor\cite{proxy_anchor} and ProxyGML\cite{proxy_gml} by more than 4.5 \%, 2.2 \% and 2.6\% in the Recall@1 (R@1) metric on the Cars-196, CUB-200 and SOP datasets, respectively. It also outperforms the current state-of-the-art, the graph-based HIST\cite{lim2022hypergraph} by 3\% and 1.4\% in terms of R@1 on the Cars-196 and CUB-200 datasets.
This shows the strength of our method in learning a low-dimensional semantic representation space.
\begin{table}[H]
    \centering
    \setlength{\tabcolsep}{1em}
     \begin{adjustbox}{width=\textwidth}
    \begin{tabular}{| l  || c  c  c  c  || c  c  c  c  |}
    \hline
    \textbf{Benchmarks $ \rightarrow$} & \multicolumn{4}{c||}{\textbf{CUB-200-2011}} & \multicolumn{4}{c|}{\textbf{Cars-196}}\\ \hline \hline
    Methods $\downarrow$   & 
        R@1 & R@2 & R@4 & R@8 &
        R@1 & R@2 & R@4 & R@8 \\ \hline   
    
    MultiSimilarity~\cite{multisimilarity_dml} &
        57.4 & 69.8 & 80.0 & 87.8 & 
        77.3 & 85.3 & 90.5 & 94.2 \\ 
    SemiHard~\cite{schroff_facenet_2015} & 
       42.6 & 55.0 & 66.4 & - & 
        51.5 & 63.8 & 73.5 & -  \\ 
     LiftedStruct~\cite{sopdataset} & 
       43.6 & 56.6 & 68.6 & 79.6 & 
        53.0 & 65.7 & 76.0 & 84.3  \\ 
    N-Pair~\cite{sohn2016improved} &
       51.0 & 63.3 & 74.3 & 83.2 & 
        71.1 & 79.7 & 86.5 & 91.6  \\
    ProxyNCA~\cite{proxy_nca} & 
       49.2 & 61.9 & 67.9 & 72.4 & 
        73.2 & 82.4 & 86.4 & 88.7  \\
    SoftTriple~\cite{soft_triplet} & 
        60.1 & 71.9 & 81.2 & 88.5 & 
        78.6 & 86.6 & 91.8 & 95.4  \\
    Clustering~\cite{clust_dml} &
        48.2 & 61.4 & 71.8 & 81.9 & 
        58.1 & 70.6 & 80.3 & 87.8  \\
    ProxyAnchor~\cite{proxy_anchor} & 
       61.7 & 73.0 & 81.8 & 88.8 &
        78.8 & 87.0 & 92.2 & 95.5  \\ 
    ProxyGML~\cite{proxy_gml} & 
       59.4 & 70.1 & 80.4 & - & 
        78.9 & 87.5 & 91.9 & -  \\ 
    HIST~\cite{lim2022hypergraph} & 
       62.5 & 73.6 & 83.0 & 89.6 & 
       80.4 & 87.6 & 92.4 & 95.4  \\ \hline \hline
    \textbf{Ours} & 
       \textbf{63.9 $\pm$ 0.3} & \textbf{74.7 $\pm$ 0.2} & \textbf{83.5 $\pm$ 0.1} & \textbf{90.1 $\pm$ 0.1} & 
       \textbf{83.4 $\pm$ 0.2} & \textbf{89.9 $\pm$ 0.2} & \textbf{94.2 $\pm$ 0.1} & \textbf{97.1 $\pm$ 0.1}  \\ 
        \hline
    \end{tabular}
    \end{adjustbox}
    \vspace{0.1mm}
    \caption{
        Comparison of the Recall@$K$ ($\%$) achieved by our method on the CUB-200-2011 and Cars-196 datasets with state-of-the-art baselines when using an embedding size of 64, showing that it outperforms all other methods. We compute recall for our method as an average over 5 runs as is done by other baselines which report this number. 
        }
    \label{tab:eval_small_embed}
    \vspace*{-4mm}
\end{table}

\begin{table}[H]
    \centering
    \setlength{\tabcolsep}{8pt}
    
    
    \begin{tabular}{|l || c  c  c  c|}
    \hline
    \multicolumn{1}{|l||}{\textbf{Benchmarks $ \rightarrow$}} & \multicolumn{4}{c|}{\textbf{SOP}} \\ \hline \hline
    Methods $\downarrow$  & 
        R@1 & R@10 & R@100 & R@1000  \\ \hline 
   MultiSimilarity~\cite{multisimilarity_dml} &
       74.1 & 87.8 & 94.7 & 98.2 \\ 
     LiftedStruct~\cite{sopdataset} & 
      62.5 & 80.8 & 91.9 & - \\ 
    N-Pair~\cite{sohn2016improved} &
       67.7 & 83.8 & 93.0 & 97.8  \\
    ProxyNCA~\cite{proxy_nca} & 
        73.7 & - & - & -  \\
    SoftTriple~\cite{soft_triplet} &   
        76.3 & 89.1 & 95.3 & -  \\
    Clustering~\cite{clust_dml} &
       67.0 & 83.7 & 93.2 & -  \\
    ProxyAnchor~\cite{proxy_anchor} & 
        76.5 & 89.0 & 95.1 & 98.2  \\ 
    ProxyGML~\cite{proxy_gml} & 
      76.2 & 89.4 & 95.4 & -  \\ 
    HIST~\cite{lim2022hypergraph} &  
      78.9 & 90.5 & 95.8 & 98.5  \\ \hline \hline
    \textbf{Ours} &  
       \textbf{79.5 $\pm$ 0.2} & \textbf{90.8 $\pm$ 0.1} & \textbf{96.3 $\pm$ 0.1} & \textbf{98.6 $\pm$ 0.1}  \\ 
        \hline

    \end{tabular}

    
    \caption{
        Comparison of the Recall@$K$ ($\%$) achieved by our method on the SOP dataset with state-of-the-art baselines when using an embedding size of 64, showing that it outperforms all other methods. We compute recall for our method as an average over 5 runs as is done by other baselines which report this number.}
    \label{tab:eval_small_sop}
    \vspace*{-4mm}
\end{table}

\subsection{Evaluation under MLRC Protocol}\label{sec:results_mlrc}
\begin{table*}[hbt!]
    \centering
     \begin{adjustbox}{width=\textwidth}
     
\begin{tabular}{|l|| c c c || c c c|}
\hline \textbf{Embedding type $ \rightarrow$} & \multicolumn{3}{c ||}{ \textbf{Concatenated (512-dim)} } & \multicolumn{3}{c | }{ \textbf{Separated (128-dim)} } \\
\hline \hline
\textbf{Methods $\downarrow$}  & \textbf{P@ 1} & \textbf{RP} & \textbf{MAP@R} & \textbf{P@ 1} & \textbf{RP} & \textbf{MAP@R} \\ \hline 
Contrastive~\cite{chopra2005learning} & $81.8 \pm 0.4$ & $35.1 \pm 0.5$ & $24.9 \pm 0.5$ & $69.8 \pm 0.4$ & $27.8 \pm 0.3$ & $17.2 \pm 0.4$ \\
Triplet~\cite{triplet_orig}& $79.1 \pm 0.4$ & $33.7 \pm 0.5$ & $23.0 \pm 0.5$ & $65.7 \pm 0.6$ & $26.7 \pm 0.4$ & $15.8 \pm 0.4$ \\
 N-Pair~\cite{sohn2016improved} & $81.0 \pm 0.5$ & $35.0 \pm 0.4$ & $24.4 \pm 0.4$ & $68.2 \pm 0.4$ & $27.7 \pm 0.2$ & $16.8 \pm 0.2$ \\
 ProxyNCA~\cite{proxy_nca} & $83.6 \pm 0.3$ & $35.6 \pm 0.3$ & $25.4 \pm 0.3$ & $73.5 \pm 0.2$ & $28.9 \pm 0.2$ & $18.3 \pm 0.2$ \\
 Margin~\cite{wu_sampling_2018} &  $81.2 \pm 0.5$ & $34.8 \pm 0.3$ & $24.2 \pm 0.3$ & $68.2 \pm 0.4$ & $27.2 \pm 0.2$ & $16.4 \pm 0.2$ \\
 Margin/class~\cite{wu_sampling_2018} & $80.0 \pm 0.6$ & $33.8 \pm 0.5$ & $23.1 \pm 0.6$ & $67.5 \pm 0.6$ & $26.7 \pm 0.4$ & $15.9 \pm 0.4$ \\
 N. Softmax~\cite{norm_softmax} &$83.2 \pm 0.3$ & $36.2 \pm 0.3$ & $26.0 \pm 0.3$ & $72.6 \pm 0.2$ & $29.3 \pm 0.2$ & $18.7 \pm 0.2$ \\
 CosFace~\cite{cosface} & $85.5 \pm 0.2$ & $37.3 \pm 0.3$ & $27.6 \pm 0.3$ & $74.7 \pm 0.2$ & $29.0 \pm 0.1$ & $18.8 \pm 0.1$ \\
ArcFace~\cite{arcface} &$85.4 \pm 0.3$ & $37.0 \pm 0.3$ & $27.2 \pm 0.3$ & $72.1 \pm 0.4$ & $27.3 \pm 0.2$ & $17.1 \pm 0.2$ \\
 FastAP~\cite{fastap} & $78.5 \pm 0.5$ & $33.6 \pm 0.5$ & $23.1 \pm 0.6$ & $65.1 \pm 0.4$ & $26.6 \pm 0.4$ & $15.9 \pm 0.3$ \\
 SNR~\cite{snr_dml} & $82.0 \pm 0.5$ & $35.2 \pm 0.4$ & $25.0 \pm 0.5$ & $69.7 \pm 0.5$ & $27.5 \pm 0.3$ & $17.1 \pm 0.3$ \\
 MultiSimilarity~\cite{multisimilarity_dml} & $85.1 \pm 0.3$ & $38.1 \pm 0.2$ & $28.1 \pm 0.2$ & $73.8 \pm 0.2$ & $29.9 \pm 0.2$ & $19.3 \pm 0.2$ \\
 MS+Miner~\cite{multisimilarity_dml} &  $83.7 \pm 0.3$ & $37.1 \pm 0.3$ & $27.0 \pm 0.4$ & $71.8 \pm 0.2$ & $29.4 \pm 0.2$ & $18.9 \pm 0.2$ \\
 SoftTriple~\cite{soft_triplet} & $84.5 \pm 0.3$ & $37.0 \pm 0.2$ & $27.1 \pm 0.2$ & $73.7 \pm 0.2$ & $29.3 \pm 0.2$ & $18.7 \pm 0.1$ \\
 ProxyAnchor~\cite{proxy_anchor} & $83.3 \pm 0.4 $ &  $ 35.7 \pm 0.3 $ & $ 25.7 \pm 0.4 $ & $ 73.7 \pm 0.4 $ & $29.4 \pm 0.3 $  & $ 18.9 \pm 0.2$ \\
  HIST~\cite{lim2022hypergraph} & $87.7 \pm 0.2$ & $ 39.9 \pm 0.2$ & $30.5 \pm 0.2$ & $ 79.3 \pm 0.2$ & $32.8 \pm 0.2$ & $22.3 \pm 0.2$  \\
  \hline \hline
 \textbf{Ours} & $ \textbf{88.4} \pm \textbf{0.2} $ &  $\textbf{ 40.1} \pm \textbf{0.2}$ & $ \textbf{31.0} \pm \textbf{0.3} $ & $ \textbf{81.2} \pm \textbf{0.2} $& $ \textbf{33.6} \pm \textbf{0.3} $ & $ \textbf{22.9} \pm \textbf{0.1} $ \\
\hline
\end{tabular}
\end{adjustbox}

\caption{ Comparison of the Precision@1, R-Precision (RP) and the Mean Average Precision @ R (MAP@R) as defined in \cite{RealityCheck} achieved by our method on the Cars-196 dataset with state-of-
the-art baselines under MLRC\cite{RealityCheck} settings. }
\label{tab:mlrc_cars}
\end{table*}

\begin{table*}[hbt!]
    \centering
     \begin{adjustbox}{width=\textwidth}
\begin{tabular}{|l|| c c c || c c c|}
\hline \textbf{Embedding type $ \rightarrow$} & \multicolumn{3}{ c || }{ \textbf{Concatenated (512-dim)} } & \multicolumn{3}{c | }{ \textbf{Separated (128-dim)} } \\
\hline \hline
\textbf{Methods $\downarrow$}  & \textbf{P@ 1} & \textbf{RP} & \textbf{MAP@R} & \textbf{P@ 1} & \textbf{RP} & \textbf{MAP@R} \\ \hline 
 Contrastive~\cite{chopra2005learning}& $68.1 \pm 0.3$ & $37.2 \pm 0.3$ & $26.5 \pm 0.3$ & $59.7 \pm 0.4$ & $32.0 \pm 0.3$ & $21.2 \pm 0.3$ \\
 Triplet~\cite{triplet_orig}& $64.2 \pm 0.3$ & $34.6 \pm 0.2$ & $23.7 \pm 0.2$ & $55.8 \pm 0.3$ & $29.6 \pm 0.2$ & $18.8 \pm 0.2$ \\
 N-Pair~\cite{sohn2016improved} & $66.6 \pm 0.3$ & $36.0 \pm 0.2$ & $25.1 \pm 0.2$ & $58.1 \pm 0.2$ & $30.8 \pm 0.2$ & $19.9 \pm 0.2$ \\
 ProxyNCA~\cite{proxy_nca} & $65.7 \pm 0.4$ & $35.1 \pm 0.3$ & $24.2 \pm 0.3$ & $57.9 \pm 0.3$ & $30.2 \pm 0.2$ & $19.3 \pm 0.2$ \\
 Margin~\cite{wu_sampling_2018} & $63.6 \pm 0.5$ & $33.9 \pm 0.3$ & $23.1 \pm 0.3$ & $54.8 \pm 0.3$ & $28.9 \pm 0.2$ & $18.1 \pm 0.2$ \\
 Margin/class~\cite{wu_sampling_2018} & $64.4 \pm 0.2$ & $34.6 \pm 0.2$ & $23.7 \pm 0.2$ & $55.6 \pm 0.2$ & $29.3 \pm 0.2$ & $18.5 \pm 0.1$ \\
 N. Softmax~\cite{norm_softmax} & $65.6 \pm 0.3$ & $36.0 \pm 0.2$ & $25.3 \pm 0.1$ & $58.8 \pm 0.2$ & $31.8 \pm 0.1$ & $21.0 \pm 0.1$ \\
 CosFace~\cite{cosface} & $67.3 \pm 0.3$ & $37.5 \pm 0.2$ & $ 26.7 \pm 0.2$ & $59.6 \pm 0.4$ & $32.0 \pm 0.2$ & $21.2 \pm 0.2$ \\
 ArcFace~\cite{arcface} & $67.5 \pm 0.3$ & $37.3 \pm 0.2$ & $26.5 \pm 0.2$ & $60.2 \pm 0.3$ & $32.4 \pm 0.2$ & $21.5 \pm 0.2$ \\
 FastAP~\cite{fastap} & $63.2 \pm 0.3$ & $34.2 \pm 0.2$ & $23.5 \pm 0.2$ & $55.6 \pm 0.3$ & $29.7 \pm 0.2$ & $19.1 \pm 0.2$ \\
 SNR~\cite{snr_dml} & $66.4 \pm 0.6$ & $36.6 \pm 0.3$ & $25.8 \pm 0.4$ & $58.1 \pm 0.4$ & $31.2 \pm 0.3$ & $20.4 \pm 0.3$ \\
 MultiSimilarity~\cite{multisimilarity_dml} & $65.0 \pm 0.3$ & $35.4 \pm 0.1$ & $24.7 \pm 0.1$ & $57.6 \pm 0.2$ & $30.8 \pm 0.1$ & $20.2 \pm 0.1$ \\
 MS+Miner~\cite{multisimilarity_dml} & $67.7 \pm 0.2$ & $37.3 \pm 0.2$ & $26.5 \pm 0.2$ & $59.4 \pm 0.3$ & $31.9 \pm 0.1$ & $21.0 \pm 0.1$ \\
 SoftTriple~\cite{soft_triplet} & $67.3 \pm 0.4$ & $37.3 \pm 0.2$ & $26.5 \pm 0.2$ & $59.9 \pm 0.3$ & $32.1 \pm 0.1$ & $21.3 \pm 0.1$ \\
 ProxyAnchor & $65.2 \pm 0.2$ & $36.0 \pm 0.2$ & $25.3 \pm 0.1$ & $56.6 \pm 0.1$ & $30.5 \pm 0.1$ & $19.8 \pm 0.2$ \\
 HIST~\cite{lim2022hypergraph} & $69.6 \pm 0.3$ & $38.8 \pm 0.1$ & $28.2 \pm 0.1$ & $61.3 \pm 0.2$ & $33.1 \pm 0.2$ & $22.3 \pm 0.1$  \\
\hline \hline
 \textbf{Ours} & $\textbf{70.1} \pm \textbf{0.2} $ &  $\textbf{39.9} \pm \textbf{0.1}$ & $\textbf{29.4} \pm \textbf{0.3}$ & $ \textbf{62.4} \pm \textbf{0.1} $ & $\textbf{33.8} \pm \textbf{0.2} $  & $ \textbf{23.1} \pm \textbf{0.3}$ \\ \hline
\end{tabular}
\end{adjustbox}
\caption{ Comparison of the Precision@1, R-Precision (RP) and the Mean Average Precision @ R (MAP@R) as defined in \cite{RealityCheck} achieved by our method on the CUB-200-2011 dataset with state-of-the-art baselines under MLRC\cite{RealityCheck} settings. }
\label{tab:mlrc_cub}
\end{table*}
\textbf{Context:} In Section 4.2, we compared the image retrieval performance of our method with other recent techniques using standard evaluation settings (backbone, embedding dimensions, image sizes) used in \cite{proxy_anchor, lim2022hypergraph, DRML, das_eccv22}. Recently, some studies \cite{RealityCheck,revisiting_dml} have pointed to flaws in these settings, including lack of a standardized backbone architecture, weakness of the metrics used, and lack of a standardized validation subset. Though we address some of these flaws by comparing against methods using the same experimental settings  as described in Section 4.1, in this section we additionally evaluate our method under the constrained protocol proposed in \cite{RealityCheck}. The constrained protocol proposes using fixed optimization settings with no learning rate scheduling to train an Inception with BatchNorm architecture. It also introduces new, more informative metrics (the R-Precision and Mean Average Precision@R). Further details of the constrained protocol can be found in ~\cite{RealityCheck}.

\textbf{Results:} We evaluate the performance of our method using models trained under the constrained protocol on the Cars-196 \cite{carsdataset} and CUB-200-2011 \cite{cubdataset} datasets. As seen in Tables \ref{tab:mlrc_cars} \ref{tab:mlrc_cub}, our method significantly outperforms all previous methods on all metrics. We significantly outperform the previous best pair-based method, the MultiSimilarity loss \cite{multisimilarity_dml} by 7.4\% and 4.8 \% in terms of P@1 (128-dim embeddings) on the Cars-196 and CUB-200 datasets respectively. We also outperform the current state-of-the-art method \cite{lim2022hypergraph}, HIST by 1.9 \% and 1.1 \% in terms of P@1 (128-dim) on the Cars-196 and CUB-200 datasets respectively. We note that these gains are higher than the improvements made by the current state-of-the-art HIST (1.5\% and 1.1 \% ) over previous methods on these benchmarks. Our method also outperforms all previous methods in terms of the R-Precision and Mean Average Precision @ R metric, demonstrating the quality of the semantic metric learned by it.

\begin{table}[H]
    \centering
    \setlength{\tabcolsep}{1em}
    \begin{tabular}{|l| c | c | c | c | c | c |}
    \hline
    \textbf{Dataset} $\rightarrow$ &\multicolumn{2}{c|}{\textbf{CUB-200-2011}} & \multicolumn{2}{c|}{\textbf{Cars-196}} & \multicolumn{2}{c|}{\textbf{SOP}} \\ \hline
    Arch/Dim & R50/512 & IBN/512 & R50/512 & IBN/512 & R50/512 & IBN/512 \\ 
    \hline 
        optimizer & Adam & Adam & Adam & Adam & Adam& Adam \\
        learning rate & $1e^{-4}$ & $1e^{-4}$ & $1e^{-4}$ & $1e^{-4}$ & $1e^{-4}$ & $1e^{-4}$\\
        weight decay & $5e^{-4}$ & $5e^{-4}$ & $1e^{-4}$ & $1e^{-4}$ & $5e^{-4}$& $5e^{-4}$\\
         Batch Size & 100 & 180 & 100 & 180 &  100 & 180\\
        BN freeze & Yes & Yes & Yes & Yes & No & No\\
        Warm-up & 1 & 1 & 1 & 1 & 0  & 0\\
        lr for proxies & 0.01 & 0.01 & 0.01 & 0.01  & 0.01 & 0.01\\ \hline
    \end{tabular}
    \caption{Hyperparameter details for Potential Field based DML for experiments described in Section 4 of the main paper.}
    \label{tab:hyper_param1}
\end{table}
\section{Hyperparameters} \label{sec:hyperparam}
For easy reproducibility, Table \ref{tab:hyper_param1} presents further details about the hyperparameters used in our experiments on all 3 datasets described in Section 4. 

\section{Computational Complexity vs Number of proxies} \label{sec:compute}
\begin{table}[H]
    \centering
    \begin{tabular}{|c|c|}
    \hline
     \textbf{ Methods} $\downarrow$   &  \textbf{Avg. time per epoch (seconds)} \\ \hline
       ProxyNCA \cite{proxy_nca} & 14.2 $\pm 0.1$ \\ \hline
       Proxy Anchor \cite{proxy_anchor} & 14.2 $\pm 0.1$  \\ \hline
       Potential Field (Ours) M=5 & 14.3$\pm 0.1$ \\ \hline
       Potential Field (Ours) M=20 & 14.5$\pm 0.1$ \\ \hline
       Potential Field (Ours) M=30 & 14.8$\pm 0.1$ \\ \hline
    \end{tabular}
    \caption{ Average time (in seconds) required to run an epoch of training on the Cars-196 dataset with a ResNet50 backbone using various methods. These results demonstrate that the time complexity of our method is similar to previous proxy-based methods. Also, note that an increase in M does not significantly alter the time complexity of our method once the number of parameters specifying proxies here (as is in most cases) is much lower than the total number of parameters (in the neural network) being trained. The standard evaluation settings of backbone, embedding dimensions, image sizes as given in Sec. 4.1 of the main paper were employed for all methods. Times were measured on a machine equipped with a single A4000 GPU over 20 epochs.   }
    \label{tab:my_label}
\end{table}
\section{Visual Results}\label{sec:visual}

\begin{figure}[H]
    \centering
    \includegraphics[width = 0.47\textwidth]{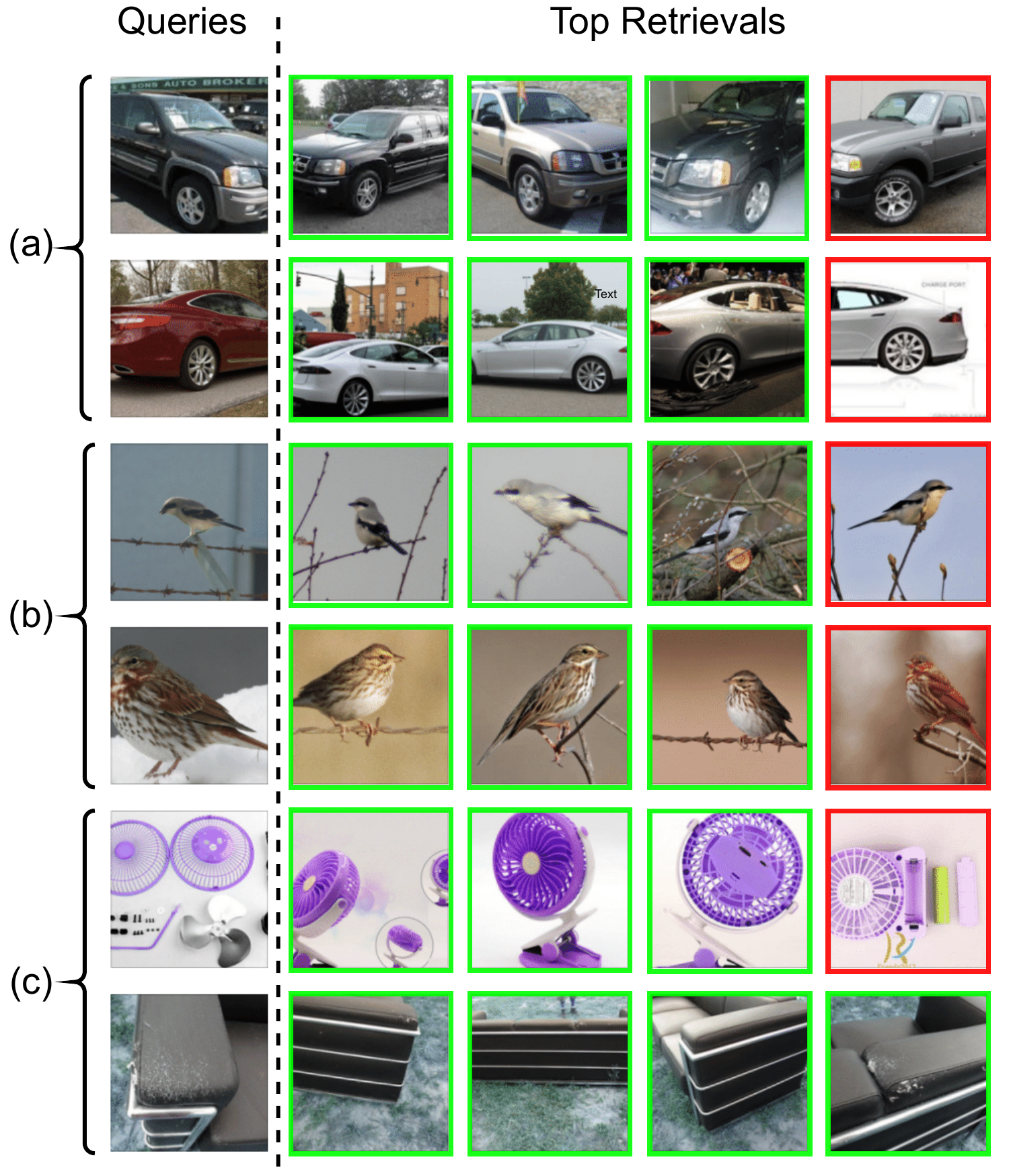}
    \caption{Example image retrieved by our method for query images from (a) Cars-196 (b) CUB-200-2011 and (c) SOP test datasets, in increasing order of distance from the query. Correct retrievals have a green border, while incorrect ones have a red one. }
    \label{fig:qualitative_res}
\end{figure}

We present qualitative results for image retrieval by our method to evaluate the semantic similarity metric learned by it. Figure \ref{fig:qualitative_res} displays 2 examples of query images from each of the 3 datasets, followed by 4 nearest images retrieved by our method, arranged in increasing order of distance. It can be seen that despite the large intra-class variation (pose, color) in the datasets, our method is able to effectively retrieve similar images. 

\begin{figure}[H]
    \centering
    \includegraphics[width = \textwidth]{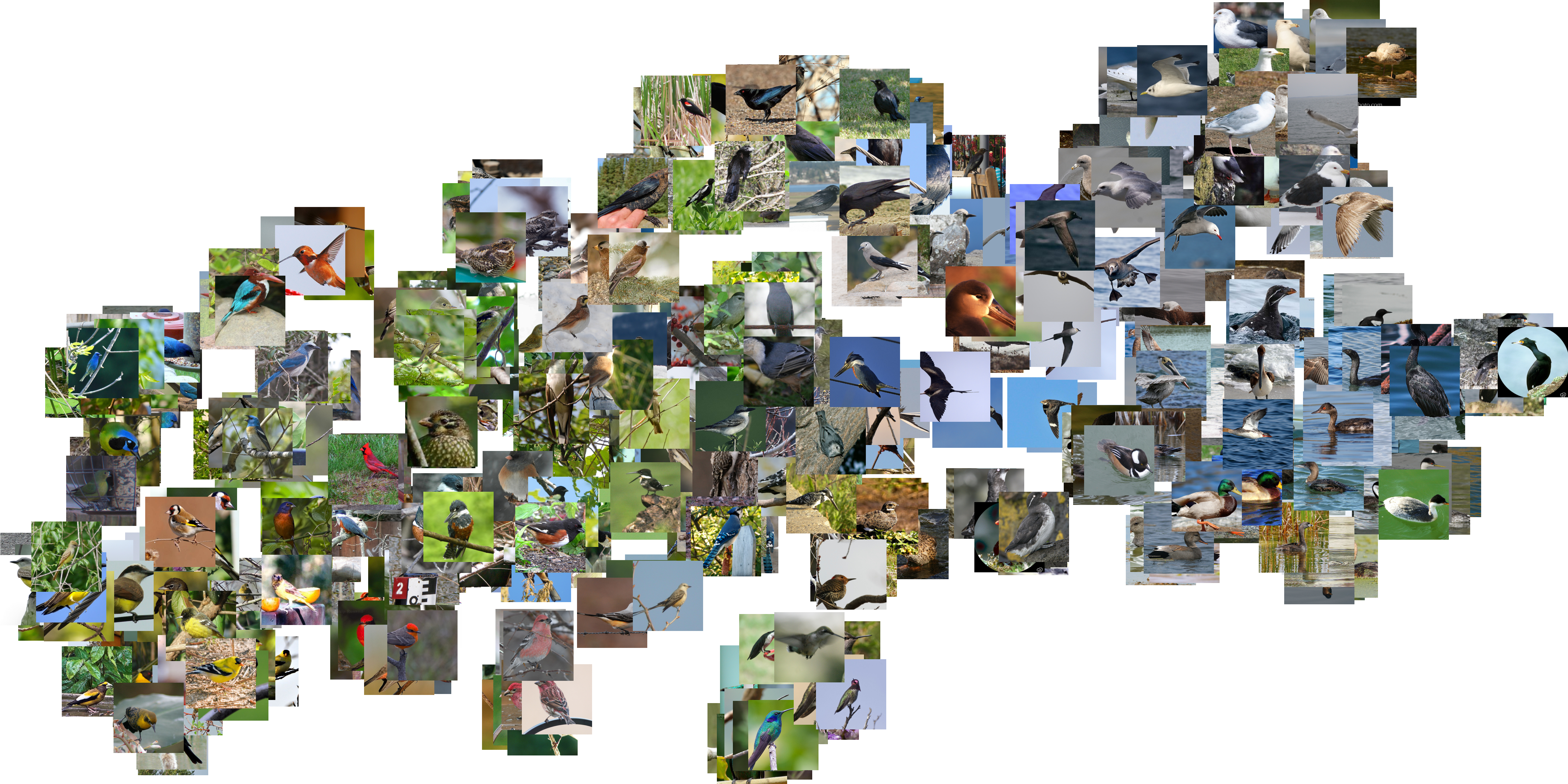}
    \caption{A t-sne visualization of a semantic representation space learnt by our method on the CUB-200 dataset }
    \label{fig:tsne_visual}
\end{figure}

Figure \ref{fig:tsne_visual} displays a t-sne visualization of the embedding space learnt by our method on the CUB-200-2011 dataset. it can be seen that images closer together share more semantic characteristics than those that are far apart.


\end{document}